\documentclass{article}
\usepackage{textcomp}

\usepackage{arxiv}

\usepackage[utf8]{inputenc} 
\usepackage[T1]{fontenc}    
\usepackage{hyperref}       
\usepackage{url}            
\usepackage{booktabs}       
\usepackage{amsfonts}       
\usepackage{nicefrac}       
\usepackage{float}          
\usepackage{microtype}      
\usepackage{lipsum}		
\usepackage{graphicx}
\usepackage{natbib}
\usepackage{doi}

\usepackage{multirow}
\usepackage{amssymb}
\usepackage{makecell}
\usepackage[table]{xcolor}
\usepackage{courier}
\usepackage{amsmath}
\usepackage{flafter}
\usepackage{wrapfig}
\usepackage{placeins}
\usepackage{dblfloatfix}

\title{EvoKnow: Continual Knowledge Evolution for AI-Generated Image Detection}

\author{
Zhiheng Peng$^{1}$,
Wenwei Jin$^{1}$\thanks{Project Lead.}
\hspace{0.15em}\thanks{Corresponding author.},
Yangshi Ge$^{1}$,
Siyu Xia$^{2}$,
Jiawei Li$^{1}$ \& Xu Tang$^{1}$ \\
$^{1}$Xiaohongshu Inc. \\
$^{2}$Southeast University \\
\texttt{\{pengzhiheng,geyangshi,wangdesheng,tangshen\}@xiaohongshu.com} \\
\texttt{\{wenwei1217.jin,siyuxia\}@gmail.com}
}

\begin{document}
\maketitle

\begin{abstract}
AI-generated image detectors are commonly trained on fixed generator
domains and become difficult to maintain as new generative models emerge.
Continual adaptation is challenging because replaying historical generated
images is costly, whereas updating shared parameters with limited
current-domain data can overwrite prior forensic knowledge. We propose
\textbf{EvoKnow}, a replay-free framework that formulates continual
AI-generated image detection as forensic knowledge evolution. EvoKnow
preserves a shared forensic basis learned from base domains, incrementally
adds isolated residual experts for complementary generator-relevant
evidence, and retrieves expertise through an Analytical Incremental
Router (AIR) updated in closed form from current-stage generated images and accumulated sufficient statistics. Experiments demonstrate effective cross-generator generalization,
few-shot expansion, and long-horizon continual adaptation. With ten
generated images per arriving generator, EvoKnow achieves 96.70\% average
accuracy on non-base GenImage generators and 94.48\% accuracy on
Chameleon without target-benchmark adaptation. Under a strict replay-free continual learning protocol, EvoKnow achieves
state-of-the-art continual learning performance, attaining 96.32\% mean
stage-wise accuracy and 4.32\% average forgetting.
\end{abstract}


\section{Introduction}
\begin{figure}[t]
    \centering
    \includegraphics[width=0.85\linewidth]{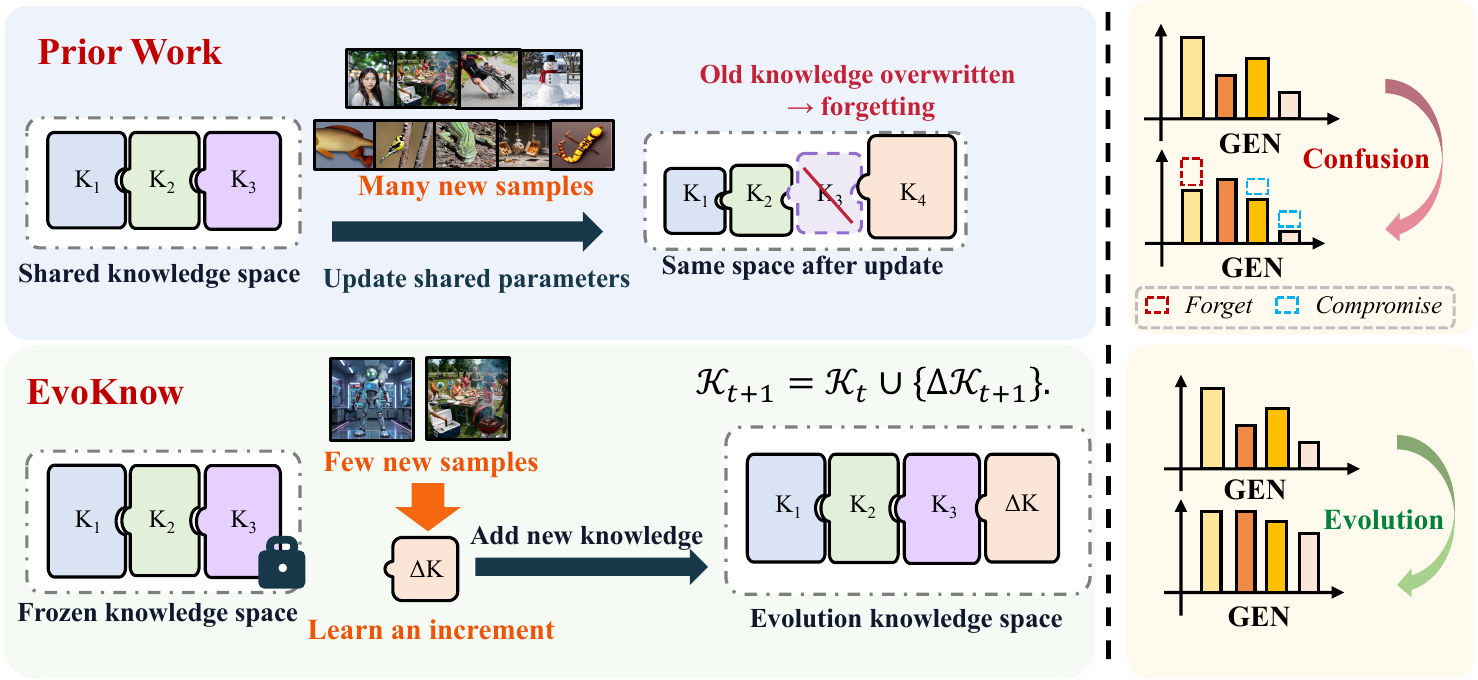}
    \caption{Continual AIGC detection with EvoKnow. Instead of repeatedly reshaping a unified forensic space, EvoKnow freezes transferable shared evidence and incrementally adds isolated residual expertise for newly arriving generators. Historical synthetic images are not replayed.}
    \label{fig:abs}
\end{figure}

Recent advances in AI image generation have substantially improved the
realism and semantic coherence of generated images. While these models
enable a wide range of creative applications, their misuse also
facilitates misinformation, identity impersonation, and forged visual
evidence. Reliable AI-generated image detection has therefore become a
fundamental problem in digital media forensics
\cite{wang2024security,zhang2025aiguard,guo2023aigc}.

Existing detectors are typically trained on data from a fixed set of
generative models and aim to generalize beyond the observed training
domains \cite{zhu2023genimage,wang2026generalizable,xu2025fully}.
However, new generative models continuously emerge and may exhibit
forensic traces absent from the original training data. When a detector
fails on an emerging generator, retraining with historical and new
generated images is costly and requires retaining an ever-growing image
collection. Alternatively, adapting a shared detector using limited
new-domain data can overwrite previously acquired forensic knowledge and
cause catastrophic forgetting.

We therefore consider \emph{generator-incremental AI-generated image
detection}, where generators arrive sequentially, only a few images are
available from each newly observed generator, and historical generated
images cannot be replayed. The objective is to expand detection coverage
to emerging generators while preserving performance on previously
observed domains. The central challenge is to adapt from limited
current-domain evidence without either overwriting prior forensic
knowledge or retaining historical generated images.

Our key observation is that forensic evidence is neither entirely shared
nor entirely domain-specific. Different generators exhibit transferable
synthetic characteristics, whereas an emerging generator may require
complementary evidence that is insufficiently captured by the existing
representation. Continually updating a single shared model promotes
transfer but is prone to forgetting, whereas learning an independent
model for every generator avoids interference but redundantly models
common evidence. As illustrated in Figure~\ref{fig:abs}, a practical
detector should instead preserve transferable forensic knowledge and
incrementally acquire complementary evidence for emerging generators.

Based on this principle, we propose \textbf{EvoKnow}
(\textbf{Evo}lution of Forensic \textbf{Know}ledge), a replay-free
knowledge-allocation framework for continual AI-generated image
detection. EvoKnow preserves a shared forensic basis learned from base
domains and frozen during incremental adaptation, while operationalizing
incremental residual knowledge through isolated, unit-specific residual
experts for arriving generator domains. By freezing the shared forensic
learner and previously learned experts, EvoKnow avoids parameter
interference during new-domain adaptation. An \emph{Analytical Incremental Router} (AIR) performs expert retrieval
through recursive closed-form updates using current-stage generated
images and accumulated sufficient statistics. Our replay-free protocol excludes retaining or replaying
historical AI-generated images and their instance-level features.
Task-agnostic real images are sampled from a predefined shared training
pool only as negative-class supervision for binary real/fake detection.

Our contributions are summarized as follows:
\begin{itemize}
    \item \textbf{A novel perspective.} We formulate continual
    AI-generated image detection as replay-free forensic knowledge
    evolution. Rather than repeatedly updating a shared detector, this
    perspective preserves transferable base-domain evidence and
    incrementally acquires complementary forensic evidence from limited
    observations of emerging generators.

    \item \textbf{A new method.} We propose \textbf{EvoKnow}, which
    combines a frozen shared forensic basis, isolated incremental residual
    experts, and AIR for analytical expert retrieval. EvoKnow expands
    detection coverage from current-stage generated images and accumulated
    router statistics without retaining historical generated images.

    \item \textbf{Strong performance.} With ten generated images per arriving generator, EvoKnow achieves 96.70\% average accuracy on non-base GenImage generators and 94.48\% accuracy on Chameleon without target-benchmark adaptation. Under a strict replay-free continual protocol, it attains 96.32\% mean stage-wise accuracy with 4.32\% average forgetting, demonstrating leading continual learning performance without historical generated-image replay.

\end{itemize}

\section{Related Work}

\subsection{Cross-Generator AI-Generated Image Detection}

Early AI-generated image detectors were often designed for specific
generative architectures or artifact patterns. More recent work instead
learns transferable forensic representations that generalize across
generators through frequency artifacts, reconstruction discrepancies,
local textures, foundation-model features, or distribution alignment
\cite{yu2025unlocking,chen2024drct,yan2024effort,tan2025c2p,yao2026safe,liu2024forgery,yang2026all}.
These methods improve detection on unseen generators by exploiting
evidence shared across generative models.

However, cross-generator detection is typically studied in a static
setting, where a detector trained on fixed source domains is expected to
generalize without further adaptation. In deployment, newly emerging
generators may exhibit forensic characteristics insufficiently covered by
the source-domain representation. Retraining on accumulated data requires
repeated optimization and storage of an ever-growing collection of
historical generated images. EvoKnow complements static cross-generator
detection by preserving a transferable shared forensic basis and
incrementally expanding detection coverage with complementary
generator-relevant evidence from newly arriving domains.

\subsection{Continual AI-Generated Image Detection}

Continual AI-generated image detection adapts a detector to sequentially
arriving generators while retaining performance on previously observed
domains. Recent studies have explored continual or parameter-efficient
formulations, including Wang et al.~\cite{wang2026generalizable},
Zhang et al.~\cite{zhang2025dynamic}, and OmniDFA
\cite{wu2025omnidfa}. More generally, continual learning mitigates
forgetting through replay, regularization, constrained optimization, or
parameter-efficient adaptation. Replay requires retaining historical
samples, whereas repeatedly updating shared parameters can still
interfere with prior knowledge.

Analytical continual learning updates regularized least-squares solutions
from accumulated sufficient statistics
\cite{zhuang2022acil,zhuang2024gacl}. Analytic subspace routing further
combines isolated low-rank experts with recursive least-squares routing
\cite{tong2025analytic}. EvoKnow adopts this analytical update principle
for AI-generated image detection, but organizes knowledge differently: it freezes a shared forensic basis and incrementally adds isolated
experts for complementary generator-relevant forensic evidence. Its Analytical Incremental Router performs expert retrieval
from current-stage generated images and accumulated sufficient
statistics, without retaining historical generated images.

\begin{figure*}[t]
\centering
\includegraphics[width=0.9\linewidth]{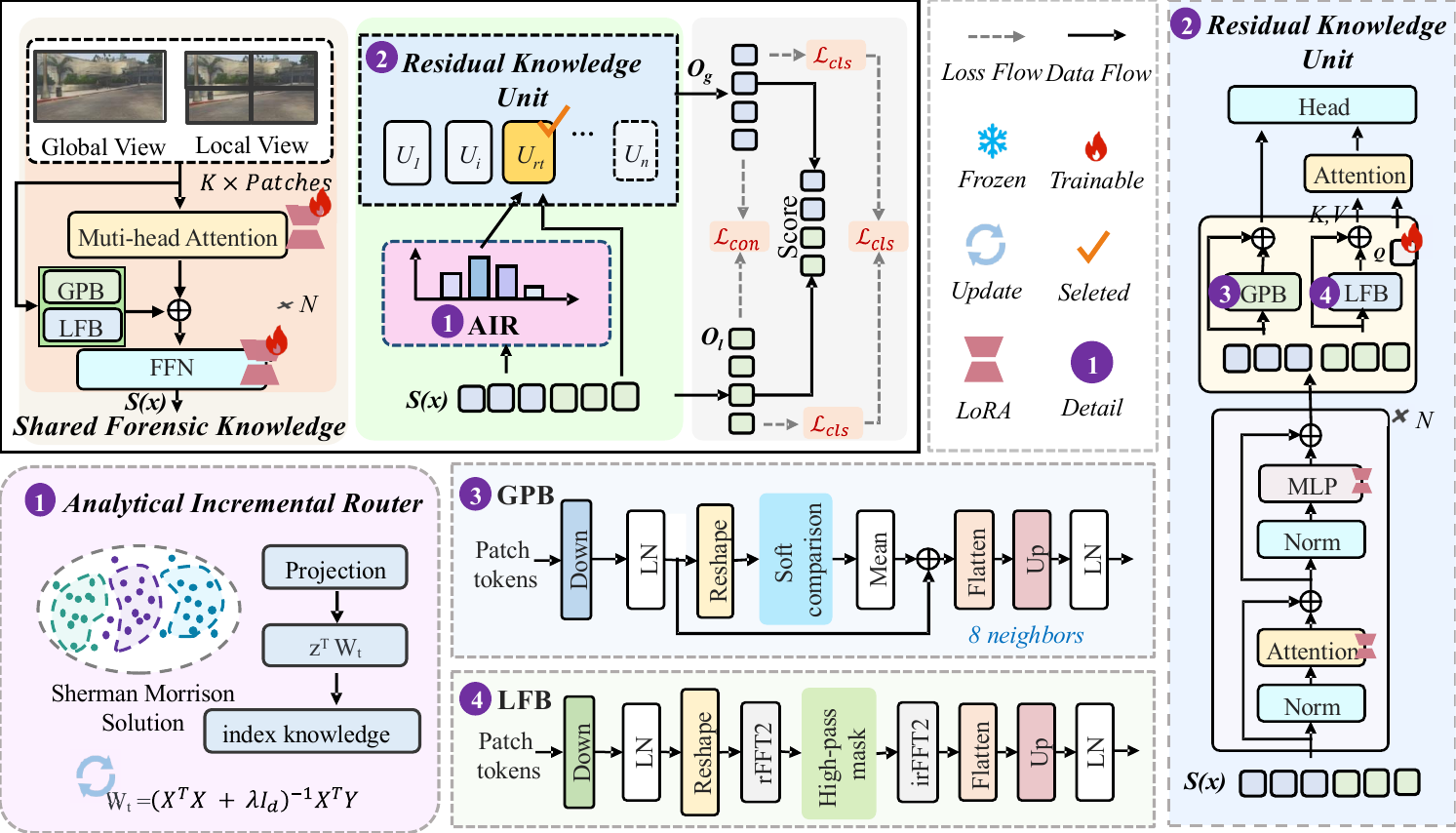}
\caption{Overview of EvoKnow. A global view and $K$ local views first
pass through a frozen shared forensic learner. AIR performs expert
retrieval over the base expert and isolated incremental residual experts.
Within each expert, LoRA adapts semantic evidence shared by both
branches, GPB emphasizes structural evidence in the global branch, and
LFB emphasizes frequency evidence in the local branch. Cross-attention
aggregates local evidence, which is fused with the global score for
real-versus-synthetic prediction. When a new generator arrives, only its
new incremental residual expert is trained and AIR is updated in closed
form; the shared forensic learner and previously learned experts remain
frozen, and historical synthetic images are not replayed.}
\label{fig:framework}
\end{figure*}

\section{Method}


As illustrated in Figure~\ref{fig:framework}, EvoKnow instantiates
continual forensic knowledge evolution through three components: a frozen
shared forensic basis that preserves transferable evidence, isolated
incremental residual experts that emphasize complementary
generator-relevant evidence from limited current-stage generated images,
and AIR for analytical expert retrieval without retaining historical
generated images. The following sections formalize the replay-free
setting and describe these components.

\subsection{Problem Setting}
We study \emph{replay-free continual AI-generated image detection} over
a sequence of emerging generator domains
$\{\mathcal{D}_t\}_{t=1}^{T}$, where each domain corresponds to a
specific generative model. At stage $t$, only a few AI-generated images
$\mathcal{D}_t^{+}$ are available. For binary real/fake training, we
randomly sample a task-agnostic real-image reference set
$\mathcal{D}_t^{-}\subset\mathcal{D}^{\mathrm{real}}$ from a predefined
shared real-image training pool. This pool is independent of generator
domains and is used only as negative-class supervision; real images are
excluded from AIR fitting. Historical generated images and their
instance-level features are unavailable and cannot be retained or
replayed.

We formulate continual adaptation as forensic knowledge evolution:
transferable evidence learned from base domains is preserved, while
complementary generator-relevant evidence is incrementally acquired. At
stage $t$, the knowledge base is
\begin{equation}
\mathcal{K}_t
=
\underbrace{\bigl\{\mathcal{K}^{\mathrm{sh}}, U_0\bigr\}}
_{\text{frozen base knowledge}}
\cup
\underbrace{\bigl\{\Delta\mathcal{K}_i\bigr\}_{i=1}^{t}}
_{\text{incremental residual knowledge}},
\end{equation}
where $\mathcal{K}^{\mathrm{sh}}$ denotes the shared forensic basis
learned by the frozen shared forensic learner, and $U_0$ denotes the
base expert trained on top of this basis using offline base-domain data.
Each $\Delta\mathcal{K}_i$ denotes generator-relevant residual knowledge
acquired for the $i$-th arriving domain and is operationalized through an
incremental residual expert $U_i$. AIR is not itself a forensic knowledge
unit; rather, it performs expert retrieval among the base expert and
incremental residual experts.

Given an image $x$, the frozen shared forensic learner outputs the
routing-cut representation $S(x)$, from which AIR derives a routing
feature $\mathbf{z}(x)=g(S(x))$. After learning stage $t$, AIR selects
from the base expert and the $t$ incremental residual experts:
\begin{equation}
r_t(x)=\mathcal{R}_t(\mathbf{z}(x))\in\{0,\ldots,t\},
\qquad
p_t(y=1\mid x)=U_{r_t(x)}(S(x)).
\end{equation}

At stage $t$, EvoKnow learns $\Delta\mathcal{K}_t$ from
$\mathcal{D}_t^{+}$ and $\mathcal{D}_t^{-}$ while freezing the shared
forensic learner, the base expert $U_0$, and all previously learned
incremental residual experts. Starting from
$\mathcal{K}_0=\{\mathcal{K}^{\mathrm{sh}},U_0\}$, the knowledge base is
updated as
\begin{equation}
\mathcal{K}_t
=
\mathcal{K}_{t-1}\cup\{\Delta\mathcal{K}_t\},
\qquad t\geq1.
\end{equation}
AIR is then updated to $\mathcal{R}_t$ using current-stage generated
images and accumulated sufficient statistics. Only learned model
parameters and aggregated router statistics are preserved across stages.

\subsection{Shared Forensic Knowledge}
The shared forensic learner is built on a pretrained image encoder, e.g.,
DINOv3, with shared LoRA modules adapted in its first $\ell$ Transformer
blocks. Given one global view $x_g$ and $K$ local views
$\{x_k^{\mathrm{loc}}\}_{k=1}^{K}$ of image $x$, the patch-token
features at the routing cut form the representation $S(x)$. This
representation provides the shared forensic basis
$\mathcal{K}^{\mathrm{sh}}$, serves as the input to the base or
incremental residual experts, and is used by AIR to derive routing
features.

The base expert $U_0$ is trained on top of
$\mathcal{K}^{\mathrm{sh}}$ using offline base-domain data. Together,
$\{\mathcal{K}^{\mathrm{sh}},U_0\}$ form the frozen base knowledge after
Phase~I. Subsequent stages therefore preserve the shared forensic basis
and adapt only through newly added incremental residual experts.

\subsection{Incremental Residual Knowledge Unit}

For an arriving generator domain, we operationalize
$\Delta\mathcal{K}_t$ through an incremental residual expert $U_t$.
Rather than explicitly modeling an additive correction to the output of
the base expert $U_0$, $U_t$ emphasizes complementary
generator-relevant forensic evidence that is insufficiently captured by
the frozen shared forensic basis. Each expert takes the routing-cut
representation $S(x)$ as input and follows the multi-view detection
paradigm of $U_0$ through isolated, unit-specific adaptation modules.

Specifically, unit-specific LoRA modules are inserted into the attention
and MLP projections of Transformer blocks following the routing cut and
are shared by the global and local branches. GPB augments the global
branch with structural evidence from differentiable neighboring-patch
comparisons, whereas LFB augments local branches with fine-grained
spectral evidence using a learnable high-pass mask. These modules produce
unit-specific residual features that are added to the corresponding
branch representations. Detailed formulations of GPB and LFB are given
in Appendix~\ref{sec:supp-gpb} and Appendix~\ref{sec:supp-lfb}.

For expert $U_i$, unit-specific heads map global and aggregated-local
features to logits $o_g^i$ and $o_l^i$, respectively. Its prediction is
\begin{equation}
U_i(S(x))
=
\left[
\operatorname{softmax}\!\left(
\operatorname{Fuse}(o_g^i,o_l^i)
\right)
\right]_{\mathrm{gen}},
\qquad i\in\{0,\ldots,t\},
\label{eq:knowledge_prediction}
\end{equation}
where $[\cdot]_{\mathrm{gen}}$ denotes the AI-generated probability.
After training, $U_t$ is frozen and added to the expert set, while the
shared forensic learner, the base expert $U_0$, and all previously
learned incremental residual experts remain frozen.

\subsection{Analytical Incremental Router}
\label{sec:router}

We introduce an \emph{Analytical Incremental Router} (AIR) for expert
retrieval. Rather than identifying the exact source generator, AIR
selects the base or incremental residual expert whose forensic evidence
is most compatible with an input. AIR derives a routing feature
$\mathbf{z}(x)=g(S(x))$ from the routing-cut representation $S(x)$,
where $g(\cdot)$ mean-pools patch tokens and applies a fixed Gaussian
projection, ReLU activation, and $\ell_2$ normalization. Since the shared
forensic learner and $g(\cdot)$ remain frozen after base training,
routing features are comparable across incremental stages.

AIR is initialized using base-domain generated images assigned to $U_0$.
When $U_t$ is added, AIR incorporates its expert-retrieval targets by
regularized least squares with recursive updates. It retains aggregated
sufficient statistics and learned router parameters, but not historical
generated images or their instance-level features. At inference, AIR
selects
\begin{equation}
r_t(x)
=
\arg\max_{0\leq i\leq t}
\mathbf{z}(x)^\top\mathbf{w}_{t,i},
\label{eq:router_selection}
\end{equation}
where $\mathbf{w}_{t,i}$ is the routing weight for $U_i$. The selected
expert then performs binary real/fake detection. Detailed AIR
initialization and recursive updates are provided in
Appendix~\ref{sec:air-details}.

\subsection{Unified Multi-view Objective and Three-phase Optimization}
\label{sec:three-phase}

\textbf{Unified Multi-view Objective.}
Both the base expert and each incremental residual expert follow the
same multi-view detection formulation. Let $p_g$, $p_l$, and $p_f$
denote predictions from the global, aggregated-local, and fused
branches, respectively. We use
\begin{equation}
\mathcal{L}_{\mathrm{cls}}
=
\sum_{b\in\{g,l,f\}}
\omega_b\operatorname{CE}(y,p_b),
\label{eq:image_classification}
\end{equation}
where $\operatorname{CE}$ denotes label-smoothed cross-entropy and
$\omega_b$ balances branch supervision. The uniformity loss
$\mathcal{L}_{\mathrm{uni}}$ encourages generated images to activate
spatially distributed regions. The weak patch-level loss
$\mathcal{L}_{\mathrm{patch}}$ encourages pooled patch responses to
remain discriminative, and the consistency loss
$\mathcal{L}_{\mathrm{cons}}$ aligns predictions across global and local
views. Their detailed definitions are provided in
Appendix~\ref{app:loss}. The unified multi-view objective is
\begin{equation}
\mathcal{L}
=
\mathcal{L}_{\mathrm{cls}}
+
\lambda_{\mathrm{uni}}\mathcal{L}_{\mathrm{uni}}
+
\lambda_{\mathrm{patch}}\mathcal{L}_{\mathrm{patch}}
+
\lambda_{\mathrm{cons}}\mathcal{L}_{\mathrm{cons}}.
\label{eq:objective}
\end{equation}

\textbf{Three-phase Optimization.}
EvoKnow evolves forensic knowledge in three phases. Phase~I learns the
shared forensic basis and base expert $U_0$ from offline base domains.
Phase~II adds an incremental residual expert for each arriving generator
domain while freezing the shared forensic learner, the base expert, and
all previously learned incremental residual experts. Phase~III expands
AIR from current-stage generated images and accumulated sufficient
statistics. Phases~I and II optimize the unified multi-view objective in
Eq.~\ref{eq:objective} over different parameter sets, whereas Phase~III
updates AIR in closed form without gradient-based optimization.

In Phase~I, we jointly optimize the shared forensic learner and base
expert, denoted by $\Theta^{\mathrm{base}}$, on offline base-domain data:
\begin{equation}
\Theta^{\mathrm{base}}
=
\arg\min_{\Theta^{\mathrm{base}}}
\mathcal{L}\big(
\mathcal{D}_{\mathrm{base}}^{+},
\mathcal{D}_{\mathrm{base}}^{-};
\Theta^{\mathrm{base}}
\big).
\end{equation}
The learned shared forensic learner and base expert $U_0$ are then
frozen.

At stage $t$ of Phase~II, we instantiate $U_t$ and optimize only its
parameters $\Theta_t$:
\begin{equation}
\Theta_t
=
\arg\min_{\Theta_t}
\mathcal{L}\big(
\mathcal{D}_t^{+},
\mathcal{D}_t^{-};
\Theta_t
\big).
\end{equation}
Only the unit-specific LoRA modules, GPB, LFB, and prediction heads of
$U_t$ are updated; the shared forensic learner, $U_0$, and
$\{U_i\}_{i=1}^{t-1}$ remain frozen. The trained $U_t$ is subsequently
frozen and added to the expert set.

In Phase~III, AIR is updated from routing features of current-stage
generated images and accumulated sufficient statistics by recursive
regularized least squares, rather than by optimizing
Eq.~\ref{eq:objective}.

\section{Experiments}

\subsection{Experimental Protocol}

\noindent\textbf{Benchmarks and Protocol.}
We evaluate EvoKnow on GenImage \cite{zhu2023genimage},
UniversalFakeDetect \cite{ojha2023towards}, and Chameleon
\cite{yan2025sanity}, covering controlled cross-generator expansion,
diverse GAN and diffusion generators, and in-the-wild images,
respectively. Unless otherwise specified, each arriving generator
provides ten AI-generated images for incremental adaptation; the
zero-shot Base model and adaptation-budget analysis are exceptions. At
each stage, a task-agnostic real-image reference set is sampled from a
predefined shared training pool for binary supervision. The pool is
disjoint from evaluation splits, independent of generator domains, and
excluded from AIR fitting. Historical generated images and their
instance-level features are neither retained nor replayed. AIR is
initialized from base-domain generated images and updated using
current-stage generated images and accumulated sufficient statistics.
Benchmark-specific sequences, base domains, and knowledge-unit
organizations are given in the corresponding settings.

\noindent\textbf{Evaluation Metrics.}
We report average accuracy (Ave.) for cross-generator and in-the-wild
evaluation. For continual detection, let
$\mathbf{B}\in\mathbb{R}^{T\times T}$ denote the task-accuracy matrix,
where $B_{i,j}$ is the accuracy on task $j$ after learning through stage
$i$, evaluated using the complete pipeline including AIR. We report the
final average accuracy and average forgetting as
\begin{equation}
\mathrm{AA}
=
\frac{1}{T}\sum_{j=1}^{T}B_{T,j},
\qquad
\mathrm{AF}
=
\frac{1}{T-1}\sum_{j=1}^{T-1}
\left(B_{j,j}-B_{T,j}\right).
\end{equation}
Higher AA and lower AF are better. For the 14-stage benchmark of Wang
et al.~\cite{wang2026generalizable}, we additionally report
$\overline{\mathrm{AA}}=\frac{1}{T}\sum_{t=1}^{T}\mathrm{AA}_t$, the
mean observed-domain accuracy over the continual trajectory.

\noindent\textbf{Baselines and Implementation.}
For cross-generator evaluation, we compare EvoKnow with detectors based
on frequency artifacts, reconstruction cues, local patterns, and
foundation-model features, including CNNSpot \cite{wang2020cnn}, UnivFD
\cite{ojha2023towards}, NPR \cite{tan2024rethinking}, FatFormer
\cite{liu2024forgery}, AIDE \cite{yan2025sanity}, SAFE
\cite{yao2026safe}, MiraGe \cite{shi2025mirage}, OmniDFA
\cite{wu2025omnidfa}, F3Net \cite{li2021frequency}, FreqNet
\cite{tan2024frequency}, C2P-CLIP \cite{tan2025c2p}, DIRE
\cite{wang2023dire}, ESSP \cite{chen2024single}, and LIDA
\cite{wang2026attribution}. These reported results follow the source-domain
protocols of their respective methods and therefore provide contextual
comparisons rather than strictly controlled same-source evaluations. For
continual detection, we compare against sequential fine-tuning, joint
training, replay-based, regularization-based, and constraint-based
methods under the reported 14-stage benchmark protocol
\cite{wang2026generalizable}. Implementation details and AIR initialization are
provided in Appendix~\ref{sec:implementation-details} and
Appendix~\ref{sec:air-details}.

\begin{table*}[!t]
\centering
\scriptsize
\setlength{\tabcolsep}{1.15pt}
\caption{
Cross-generator and cross-benchmark detection accuracy (\%).
Columns are grouped by evaluation setting.
\textbf{Cross-generator Generalization} compares existing detectors with
EvoKnow (\textbf{Base}), the zero-shot shared forensic learner and base expert.
\textbf{After 10-shot Expansion} reports EvoKnow (\textbf{Exp.}) on the
generators covered by incremental adaptation and EvoKnow (\textbf{Transfer})
on external benchmarks without target-benchmark adaptation.
Each arriving generator provides ten generated images for expansion.
Training sources and protocols are specified in the corresponding
experimental settings. For Chameleon, the base-training domain is indicated
in parentheses. Gray rows denote base domains and orange rows denote
summary metrics; base domains are excluded from averages.
\emph{Ave.} averages non-base generators, while GAN Ave. and Diffusion Ave.
summarize the corresponding UniversalFakeDetect groups.
EvoKnow columns are shaded in blue. For each average or Chameleon
accuracy row, the best and second-best values across all methods and
configurations are shown in \textbf{bold} and \underline{underlined},
respectively.
$^{\dagger}$ denotes OmniDFA models trained with UniversalFakeDetect data.
}
\label{tab:generalization}
\resizebox{\textwidth}{!}{%
\begin{tabular}{
    ll
    *{9}{c}
    >{\columncolor{blue!5}}c
    >{\columncolor{blue!5}\hspace{6pt}}c
    >{\columncolor{blue!5}}c
}
\toprule
\multirow{2}{*}{Dataset}
& \multirow{2}{*}{Generator / Metric}
& \multicolumn{10}{c}{\textbf{Cross-generator Generalization}}
& \multicolumn{2}{c}{
    \cellcolor{blue!8}\shortstack[c]{\textbf{After 10-shot}\\\textbf{Expansion}}
  } \\
\cmidrule(lr){3-12}\cmidrule(lr){13-14}
&
& CNNSpot & UnivFD & NPR & FatFormer & C2P-CLIP
& AIDE & SAFE & MiraGe & OmniDFA
& \shortstack[c]{\textbf{EvoKnow}\\\textbf{(Base)}}
& \shortstack[c]{\textbf{EvoKnow}\\\textbf{(Exp.)}}
& \shortstack[c]{\textbf{EvoKnow}\\\textbf{(Transfer)}} \\
\midrule

\rowcolor{gray!12}
& SD v1.4 \scriptsize (base)
& 96.3 & 55.6 & 55.1 & 53.6 & 77.5 & 77.2
& 99.9 & 98.8 & 97.47$^{\dagger}$
& 99.70 & 98.18 & -- \\

& BigGAN
& 46.8 & 84.4 & 57.7 & 82.2 & 85.9 & 50.6
& 77.4 & 96.5 & 97.33$^{\dagger}$
& 93.80 & 93.77 & -- \\

& Midjourney
& 52.8 & 55.1 & 53.4 & 52.1 & 56.6 & 58.2
& 95.7 & 83.2 & 97.58$^{\dagger}$
& 93.10 & 96.06 & -- \\

& SD v1.5
& 95.9 & 55.7 & 55.0 & 53.8 & 76.9 & 77.4
& 99.8 & 98.5 & 97.75$^{\dagger}$
& 99.70 & 98.53 & -- \\

& ADM
& 50.1 & 62.5 & 43.8 & 61.4 & 71.6 & 50.4
& 59.5 & 82.7 & 85.50$^{\dagger}$
& 78.10 & 96.21 & -- \\

& GLIDE
& 39.8 & 61.3 & 41.2 & 65.5 & 73.5 & 54.6
& 91.7 & 91.3 & 96.71$^{\dagger}$
& 95.50 & 96.27 & -- \\

& Wukong
& 78.6 & 61.1 & 57.4 & 60.9 & 79.4 & 70.5
& 95.7 & 97.6 & 97.78$^{\dagger}$
& 99.70 & 98.64 & -- \\

& VQDM
& 53.4 & 76.9 & 48.4 & 72.5 & 73.7 & 50.8
& 98.0 & 92.4 & 96.78$^{\dagger}$
& 95.70 & 97.41 & -- \\

\cmidrule(lr){2-14}
\rowcolor{orange!10}
\multirow{-9}{*}{GenImage}
& \textit{Ave. (excl. base)}
& 59.63 & 65.29 & 50.99 & 64.06 & 73.94 & 58.93
& 88.26 & 91.74 & \underline{95.63}$^{\dagger}$
& \cellcolor{orange!10}93.66
& \cellcolor{orange!10}\textbf{96.70}
& \cellcolor{orange!10}-- \\

\midrule

& \multicolumn{13}{l}{\textit{GAN family}} \\

\rowcolor{gray!12}
& ProGAN \scriptsize (base)
& 99.99 & 100.00 & 99.84 & 99.89 & 99.98
& 99.99 & 99.86 & 100.00 & --
& 99.98 & 100.00 & 96.50 \\

& CycleGAN
& 85.20 & 98.50 & 95.00 & 99.32 & 97.31
& 98.49 & 98.87 & 94.30 & --
& 99.36 & 99.21 & 94.80 \\

& BigGAN
& 70.20 & 94.50 & 87.55 & 99.50 & 99.12
& 83.95 & 89.73 & 96.50 & --
& 99.17 & 99.55 & 99.00 \\

& StyleGAN
& 85.70 & 82.00 & 96.23 & 97.15 & 96.44
& 99.65 & 98.04 & 96.80 & --
& 96.52 & 98.71 & 96.20 \\

& GauGAN
& 78.95 & 99.50 & 86.57 & 99.41 & 99.17
& 73.24 & 91.52 & 93.60 & --
& 99.95 & 99.88 & 98.80 \\

& StarGAN
& 91.70 & 97.00 & 99.75 & 99.75 & 99.60
& 99.90 & 99.90 & 96.10 & --
& 97.18 & 98.85 & 98.00 \\

\cmidrule(lr){2-14}
\rowcolor{orange!10}
& \textit{GAN Ave. (excl. base)}
& 82.35 & 94.30 & 93.02 & \underline{99.03} & 98.33
& 91.05 & 95.61 & 95.46 & --
& \cellcolor{orange!10}98.44
& \cellcolor{orange!10}\textbf{99.24}
& \cellcolor{orange!10}97.36 \\

\cmidrule(lr){2-14}
& \multicolumn{13}{l}{\textit{Diffusion family}} \\

& Guided
& 65.66 & 78.14 & 97.35 & 94.20 & 96.10
& 97.85 & 97.25 & 98.00 & --
& 82.75 & 87.90 & 99.10 \\

& LDM$_{200}$
& 60.07 & 70.03 & 84.55 & 76.00 & 69.10
& 88.50 & 82.35 & 82.00 & --
& 99.30 & 99.80 & 91.50 \\

& LDM$_{200c}$
& 54.03 & 94.19 & 97.65 & 98.60 & 99.25
& 98.20 & 98.80 & 98.30 & --
& 97.67 & 99.53 & 99.55 \\

& GLIDE$_{100}$
& 54.96 & 73.76 & 98.00 & 94.90 & 97.25
& 97.45 & 98.65 & 94.60 & --
& 97.42 & 98.33 & 99.10 \\

& GLIDE$_{100/27}$
& 54.14 & 94.36 & 98.20 & 98.65 & 99.30
& 98.40 & 98.80 & 98.60 & --
& 96.07 & 98.90 & 99.60 \\

& GLIDE$_{50/27}$
& 60.78 & 79.07 & 96.25 & 94.35 & 95.25
& 98.15 & 95.75 & 97.50 & --
& 96.20 & 99.03 & 98.80 \\

& GLIDE$_{100/10}$
& 63.80 & 79.85 & 97.15 & 94.65 & 95.25
& 98.40 & 96.60 & 97.50 & --
& 95.77 & 98.90 & 98.80 \\

& DALL-E
& 55.58 & 86.78 & 87.15 & 98.75 & 98.55
& 97.50 & 97.50 & 98.60 & --
& 98.77 & 99.53 & 99.20 \\

\cmidrule(lr){2-14}
\rowcolor{orange!10}
& \textit{Diffusion Ave.}
& 58.63 & 82.02 & 94.54 & 93.76 & 93.76
& 96.81 & 95.71 & 95.64 & --
& \cellcolor{orange!10}95.49
& \cellcolor{orange!10}\underline{97.74}
& \cellcolor{orange!10}\textbf{98.21} \\

\cmidrule(lr){2-14}
\rowcolor{orange!10}
\multirow{-19}{*}{\shortstack[c]{Universal\\FakeDetect}}
& \textit{Ave. (excl. base)}
& 67.75 & 86.74 & 93.95 & 95.79 & 95.51
& 94.59 & 95.67 & 95.57 & --
& \cellcolor{orange!10}96.63
& \cellcolor{orange!10}\textbf{98.32}
& \cellcolor{orange!10}\underline{97.88} \\

\midrule
\rowcolor{orange!10}
Chameleon
& \textit{Acc. (ProGAN base)}
& 56.94 & 57.22 & 57.29 & -- & -- & 58.37 & -- & -- & --
& \cellcolor{orange!10}\underline{81.89}
& \cellcolor{orange!10}--
& \cellcolor{orange!10}\textbf{88.71} \\

\rowcolor{orange!10}
Chameleon
& \textit{Acc. (SD v1.4 base)}
& 60.11 & 55.62 & 58.13 & 51.20 & 51.10
& 62.60 & -- & -- & 83.48
& \cellcolor{orange!10}\underline{93.92}
& \cellcolor{orange!10}--
& \cellcolor{orange!10}\textbf{94.48} \\

\bottomrule
\end{tabular}
}
\end{table*}

\begin{table*}[!tb]
\centering
\scriptsize
\renewcommand{\arraystretch}{1.0}
\setlength{\tabcolsep}{2.0pt}
\caption{
Continual detection on the 14-stage multi-source benchmark of Wang
et al.~\cite{wang2026generalizable}, covering GANs, deepfakes, and diffusion models.
ProGAN is the base domain, and each subsequent generator provides ten
generated images for adaptation. Baselines follow their reported
continual-learning protocols; Joint explicitly uses accumulated
historical data. In contrast, EvoKnow neither retains nor replays
historical generated images or their instance-level features, and uses
task-agnostic real-image supervision for binary detection. The first 14
columns report stage-wise observed-domain accuracy $\mathrm{AA}_t$ using
the complete AIR pipeline; $\overline{\mathrm{AA}}$ averages
$\mathrm{AA}_t$ over the trajectory, and AF is in percentage points.
Baseline AF signs are converted for consistent comparison. Best non-Joint
results are bolded, with ties highlighted once.
}
\label{tab:continual}
\resizebox{\textwidth}{!}{%
\begin{tabular}{
l
*{14}{c}
>{\columncolor{blue!10}}c
>{\columncolor{blue!10}}c
}
\toprule
\multirow{2}{*}{Method}
& \multicolumn{14}{c}{Observed-domain average accuracy
($\mathrm{AA}_t$, \%)}
& \multicolumn{2}{c}{
    \cellcolor{blue!10}\textbf{Continual summary}
  } \\
\cmidrule(lr){2-15}\cmidrule(lr){16-17}
& ProGAN
& Deepfake
& BigGAN
& StyleGAN2
& DDPM
& ADM
& DALL-E
& GLIDE
& SD v1.4
& Midjourney
& VQDM
& SD v2.1
& SDXL v1.0
& SD v3.0
& $\overline{\mathrm{AA}}\uparrow$
& $\mathrm{AF}\downarrow$ \\
\midrule
Seq
& \textbf{99.99} & 94.46 & 81.01 & 94.05 & 90.35
& 80.14 & 79.66 & 76.68 & 84.50 & 73.82
& 74.31 & 51.90 & 57.29 & 73.85
& 79.43
& 23.04 \\
\midrule
Joint
& 99.99 & 98.77 & 99.34 & 99.57 & 98.15
& 98.97 & 98.81 & 98.54 & 98.80 & 98.30
& 98.38 & 95.69 & 95.17 & 96.01
& 98.18
& 1.77 \\
\midrule
ER~\cite{chaudhry2019tiny}
& 99.99 & 98.26 & 96.77 & 96.96 & 90.92
& 85.97 & 86.18 & 84.65 & 84.11 & 83.94
& 83.73 & 84.51 & 86.02 & 90.68
& 89.48
& 12.10 \\
EWC~\cite{kirkpatrick2017overcoming}
& 99.99 & 92.57 & 91.63 & 93.73 & 89.44
& 82.87 & 79.85 & 70.47 & 82.35 & 80.11
& 77.10 & 67.97 & 69.39 & 78.11
& 82.54
& 21.52 \\
OSLA~\cite{ritter2018online}
& 99.99 & 98.37 & 90.25 & 94.98 & 87.69
& 76.65 & 85.05 & 75.53 & 83.32 & 77.86
& 85.92 & 69.61 & 73.33 & 79.08
& 84.12
& 23.77 \\
A-GEM~\cite{chaudhry2018efficient}
& 99.99 & 87.70 & 85.48 & 93.03 & 92.70
& 85.18 & 80.14 & 70.58 & 83.37 & 78.33
& 75.76 & 66.07 & 68.50 & 73.91
& 81.48
& 25.52 \\
SI~\cite{zenke2017continual}
& 99.99 & 93.82 & 82.67 & 92.32 & 90.68
& 84.65 & 76.98 & 74.66 & 80.73 & 77.57
& 80.98 & 64.58 & 69.54 & 75.26
& 81.75
& 17.40 \\
iCaRL~\cite{rebuffi2017icarl}
& 99.99 & 94.38 & 83.19 & 94.63 & 91.52
& 79.87 & 87.93 & 78.18 & 84.64 & 76.18
& 79.38 & 64.26 & 65.29 & 70.51
& 82.14
& 24.72 \\
Wang et al.~\cite{wang2026generalizable}
& 99.99 & \textbf{99.06} & \textbf{99.13}
& \textbf{99.07} & 98.38
& \textbf{97.30} & 96.13 & 95.95 & 94.90 & 93.53
& \textbf{96.20} & 87.41 & 89.55 & 92.32
& 95.64
& 8.91 \\
\midrule
\rowcolor{blue!5}
\textbf{EvoKnow (10-shot)}
& 99.97 & 95.55 & 96.75 & 94.85 & \textbf{98.81}
& 96.24 & \textbf{98.39} & \textbf{96.30}
& \textbf{98.16} & \textbf{97.39}
& 96.08 & \textbf{93.28} & \textbf{92.40}
& \textbf{94.31}
& \textbf{96.32}
& \textbf{4.32} \\
\bottomrule
\end{tabular}%
}
\end{table*}

\subsection{Generalization of Shared Forensic Knowledge}
We first evaluate whether the shared forensic basis learned in Phase~I
generalizes beyond its base domain before incremental expansion.
Table~\ref{tab:generalization} reports the zero-shot \textbf{Base}
configuration, together with the expanded configurations used in
subsequent analyses. The base-training domain is GenImage SD v1.4
for GenImage evaluation and ProGAN for UniversalFakeDetect evaluation.
Under these respective single-domain initializations, EvoKnow (Base)
achieves 93.66\% average accuracy on non-base GenImage generators and
96.63\% on UniversalFakeDetect excluding the ProGAN base domain.
On the in-the-wild Chameleon benchmark, it reaches 81.89\% accuracy
with ProGAN initialization and 93.92\% with SD v1.4 initialization.
Under the respective reported source-domain protocols in
Table~\ref{tab:generalization}, the latter result is 10.44 percentage
points higher than the reported OmniDFA result \cite{wu2025omnidfa}.
These results suggest that the frozen shared forensic basis captures
transferable real-versus-generated evidence and provides a strong
starting point for incremental knowledge expansion.

\subsection{Few-shot Incremental Knowledge Expansion and Transfer}
We next evaluate whether Phase~II expands the frozen shared forensic
basis using limited evidence from emerging generators. On GenImage, each
arriving generator adds an incremental residual expert trained with only
ten AI-generated images, while the shared forensic learner, base expert,
and previously learned experts remain frozen. Under this generator-based
10-shot setting, average accuracy on non-base generators improves from
93.66\% to 96.70\%. Notably, ADM accuracy increases from 78.10\% to
96.21\%. On UniversalFakeDetect, the corresponding ProGAN-initialized
configuration improves from 96.63\% to 98.32\% after few-shot expansion.
For external evaluation without target-benchmark adaptation, the
GenImage-expanded model achieves 97.88\% average accuracy on
UniversalFakeDetect and improves Chameleon accuracy from 93.92\%
to 94.48\%. With ProGAN initialization, expansion improves Chameleon
accuracy from 81.89\% to 88.71\%, also without target-benchmark
adaptation. These results suggest that incremental residual experts
provide complementary generator-relevant forensic evidence beyond the
frozen shared basis, improving both observed-domain coverage and external
transfer. Alternative knowledge-unit organizations are analyzed in
Section~\ref{sec:ablation}.

\subsection{Replay-free Continual Knowledge Evolution}
\label{sec:continual}

We next evaluate long-horizon continual adaptation on the 14-stage
multi-source benchmark of Wang et al.~\cite{wang2026generalizable}, spanning GANs,
deepfakes, and diffusion models. Unlike GenImage experiments, this
benchmark uses ProGAN as the base domain to initialize the shared
forensic learner, base expert $U_0$, and AIR expert-$0$ routing target.
Each subsequent generator provides ten AI-generated images, while
historical generated images are neither retained nor replayed.

Table~\ref{tab:continual} reports stage-wise observed-domain accuracy
$\mathrm{AA}_t$, its trajectory average
$\overline{\mathrm{AA}}$, and average forgetting (AF). EvoKnow achieves
96.32\% mean stage-wise accuracy with 4.32\% average forgetting, yielding
the highest reported mean stage-wise accuracy and the lowest reported
average forgetting among non-Joint methods. Unlike the Joint reference,
EvoKnow never accesses accumulated historical generated images. These
results support the knowledge-evolution design: the frozen shared
forensic learner preserves transferable evidence, incremental residual
experts add complementary evidence without modifying prior experts, and
AIR expands expert retrieval through accumulated sufficient statistics.


\subsection{Ablation and Analysis}
\label{sec:ablation}

We analyze the multi-view forensic formulation, routing representation,
and adaptation-sample budget of EvoKnow. Additional analysis of residual
knowledge organization is provided in Appendix~\ref{sec:supp-organization}.

\begin{table*}[!t]
\centering

\begin{minipage}[t]{0.53\textwidth}
\centering
\scriptsize
\setlength{\tabcolsep}{2.5pt}
\renewcommand{\arraystretch}{1.08}
\textbf{(a) GenImage ablation (mean accuracy, \%)}\\[3pt]
\resizebox{\linewidth}{!}{%
\begin{tabular}{lcccccc}
\toprule
Views & GPB & LFB & Cons. & Patch. & Base & Exp. \\
\midrule
Global
& -- & -- & -- & --
& 88.99 & 89.04 \\
Global
& \checkmark & -- & -- & --
& 89.16 & 90.32 \\
Global
& -- & \checkmark & -- & --
& 89.03 & 92.14 \\
Global
& \checkmark & \checkmark & -- & --
& 89.18 & 93.31 \\
Global+local
& \checkmark & \checkmark & -- & --
& 93.54 & 94.64 \\
Global+local
& \checkmark & \checkmark & \checkmark & --
& 93.98 & 95.87 \\
Global+local
& \checkmark & \checkmark & -- & \checkmark
& 93.87 & 95.63 \\
\rowcolor{blue!5}
Global+local
& \checkmark & \checkmark & \checkmark & \checkmark
& \textbf{94.41} & \textbf{96.88} \\
\bottomrule
\end{tabular}
}
\end{minipage}\hfill
\begin{minipage}[t]{0.45\textwidth}
\centering
\scriptsize
\setlength{\tabcolsep}{4pt}
\renewcommand{\arraystretch}{1.08}
\textbf{(b) Generator classification probes (\%)}\\[3pt]
\begin{tabular}{lcc}
\toprule
Representation & Linear probe & $k$-NN \\
\midrule
Shared & 68.58 & 67.83 \\
Shared + GPB & 69.50 & 67.92 \\
Shared + GPB + LFB & 69.50 & 68.42 \\
\rowcolor{blue!5}
Routing features & \textbf{79.42} & \textbf{78.83} \\
\bottomrule
\end{tabular}
\end{minipage}

\caption{
Detection ablation and generator-separability analysis of EvoKnow.
(a) Mean detection accuracy over all eight GenImage domains,
including the SD v1.4 base domain.
Base and Exp. denote the EvoKnow base-model generalization
and 10-shot expansion configurations in Table~\ref{tab:generalization}.
GPB and LFB denote the Global Pattern-aware Block and Local
Frequency-aware Block. Cons. and Patch. denote cross-view consistency
and weak patch-level supervision, respectively.
$\checkmark$ indicates an enabled component and -- indicates a disabled
component.
(b) Five-fold cross-validation accuracy for 14-way generator
classification using linear and $k$-NN probes on frozen representations.
Each fold uses an 80/20 training--validation split.
Routing features denote the representation supplied to AIR.
}
\label{tab:diagnostics}
\end{table*}

\noindent\textbf{Multi-view Forensic Modeling.}
Table~\ref{tab:diagnostics}(a) evaluates the contributions of view
configurations, GPB and LFB, and auxiliary supervision under both the
\textbf{Base} and \textbf{Exp.} settings. Combining global and local
views increases mean accuracy from 89.18\% to 93.54\% for Base and
from 93.31\% to 94.64\% for Exp., demonstrating the complementary
value of image-level context and local forensic details.
Adding cross-view consistency and weak patch-level supervision
further improves the full model to 94.41\% and 96.88\%, respectively.
These gains highlight the contribution of multi-view features and
complementary supervision to both shared forensic knowledge learning
and few-shot incremental expansion.

\begin{figure*}[!t]
\centering
\resizebox{\linewidth}{!}{%
    \begin{tabular}{@{}c@{\hspace{12pt}}c@{}}
        \includegraphics[height=4.2cm]
        {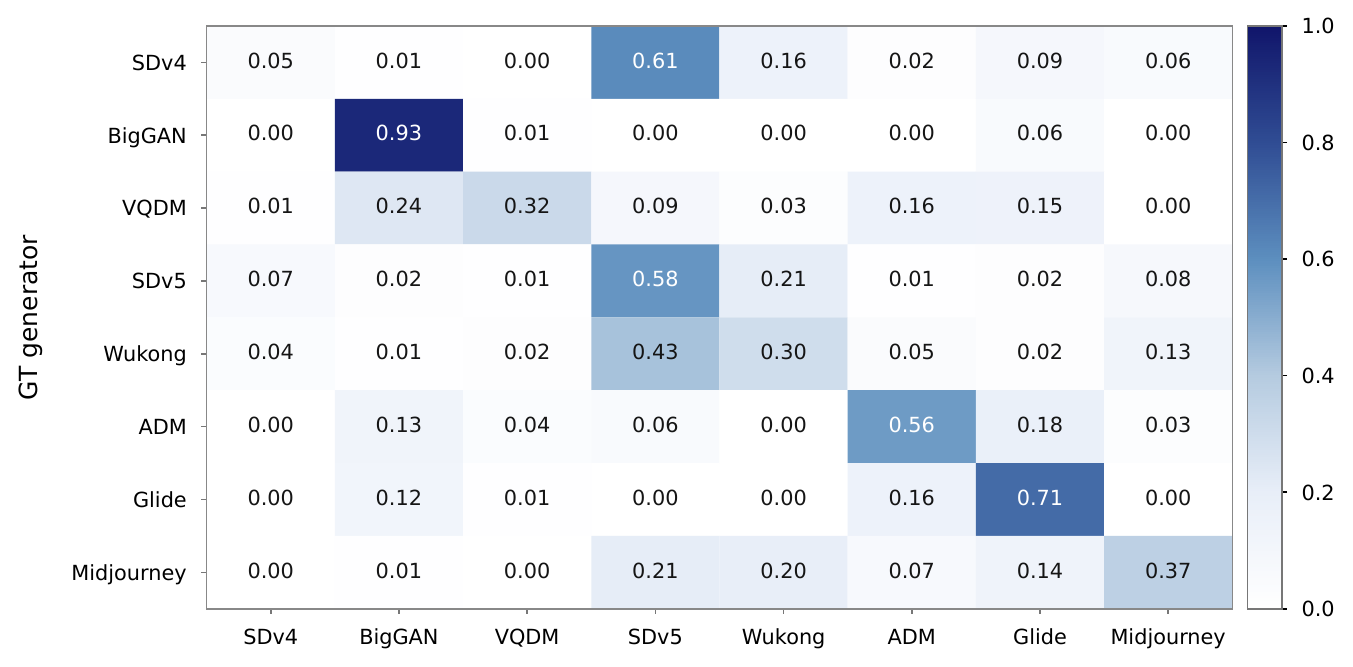}
        &
        \includegraphics[height=4.2cm]
        {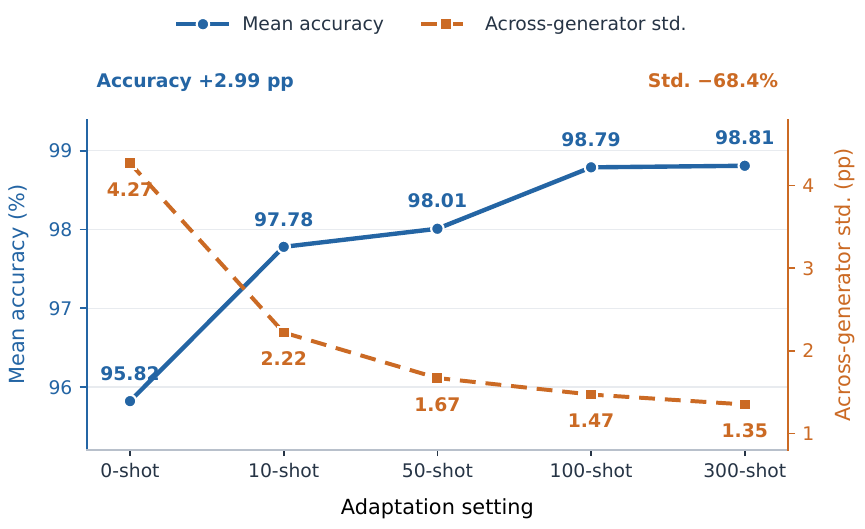}
        \\[5pt]
        {\scriptsize (a) Generator-to-expert routing assignments}
        &
        {\scriptsize (b) Accuracy and standard deviation trends}
    \end{tabular}%
}
\caption{
Expert retrieval and adaptation-sample analysis. (a) Generator-to-expert
routing assignments. (b) Effect of the GenImage adaptation budget on
cross-generator evaluation over UniversalFakeDetect, reported by average
accuracy and across-generator standard deviation. UniversalFakeDetect is
used only for evaluation.
}
\label{fig:adaptation_routing}
\end{figure*}

\noindent\textbf{Generator Separability and Expert Retrieval.}
We assess generator separability by applying linear and $k$-NN probes
to frozen features extracted at different stages of the shared forensic
learner. The training features are used to fit the linear classifier
and provide the reference set for $k$-NN.
Table~\ref{tab:diagnostics}(b) shows that routing features improve
linear-probe accuracy from 68.58\% to 79.42\% and $k$-NN accuracy from
67.83\% to 78.83\%, indicating that generator-discriminative information
is more accessible in the routing feature space.
These results support the choice of representation for expert retrieval.
Figure~\ref{fig:adaptation_routing}(a) further shows generator-dependent
expert preferences and cross-generator reuse of experts.
AIR uses the retrieved experts for binary real/fake detection,
and Appendix~\ref{sec:router-state} provides an auxiliary analysis
of how the complete detection-and-routing pipeline maps its predictions
to source labels.

\noindent\textbf{Adaptation-Sample Budget.}
Figure~\ref{fig:adaptation_routing}(b) varies the number of generated
adaptation images in the fixed GenImage stream and evaluates transfer on
UniversalFakeDetect. Larger budgets improve average accuracy and reduce
across-generator variation. Notably, ten images per arriving generator
already improve coverage beyond the frozen shared forensic basis, whereas
additional samples mainly benefit generators that are challenging under
zero-shot transfer. Complete generator-wise results are reported in
Appendix~\ref{sec:supp-fewshot}.

\begin{figure*}[t]
    \centering
    \includegraphics[width=0.75\textwidth]{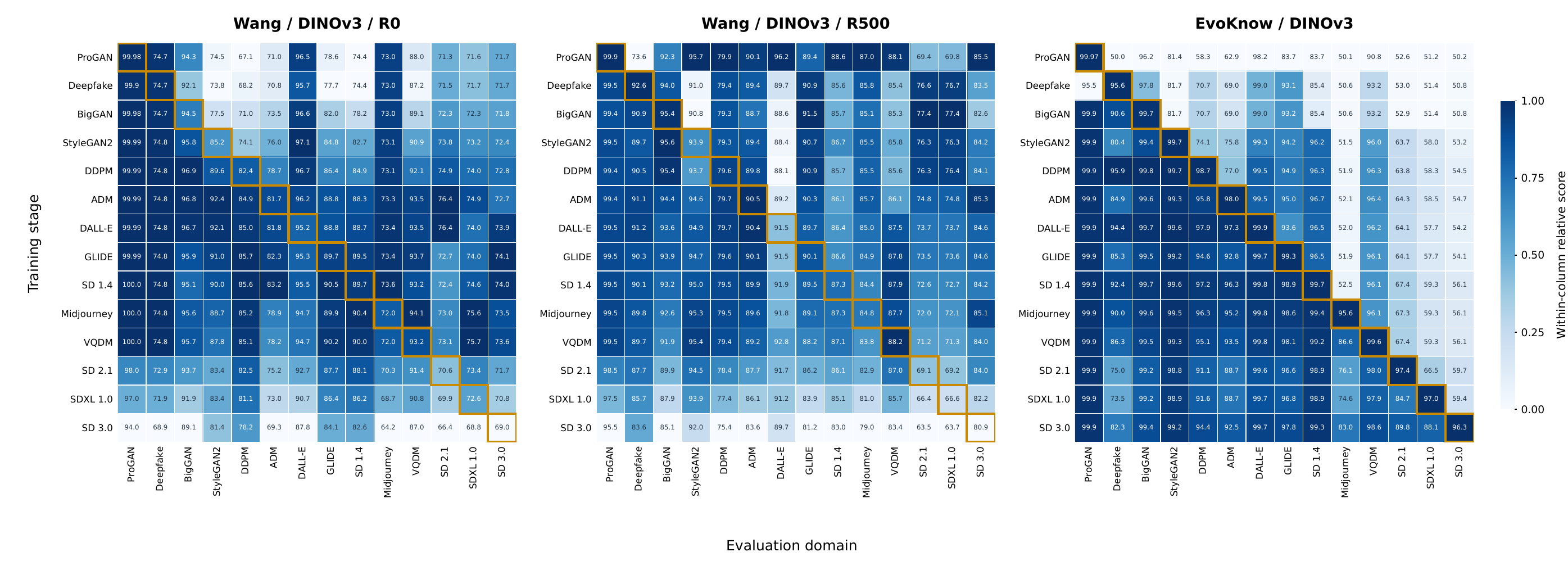}
    \caption{
    Stage-wise test accuracy and knowledge retention under the
    14-generator, 10-shot continual-learning protocol with DINOv3.
    EvoKnow is compared with Wang et al. without replay (R0)
    and with a 500-sample replay buffer (R500).
    All configurations use ten new generated images per
    incremental stage; R500 additionally uses historical samples
    from its replay buffer.
    Each entry reports test accuracy (\%) on the column-indexed
    evaluation domain after completing the row-indexed training
    task. Orange-bordered diagonal entries indicate accuracy
    immediately after task acquisition; entries below the
    diagonal track subsequent retention.
    Colors are independently min--max normalized across training
    stages for each method--domain pair, with darker blue
    indicating higher relative accuracy.
    Numerical annotations report the original accuracy values.
    }
    \label{fig:forgetting_heatmap}
\end{figure*}

\noindent\textbf{Few-shot Incremental Learning and Knowledge Retention.}
To examine detection performance and forgetting under a common
10-shot budget for new generators, we compare EvoKnow with
Wang et al. in both R0 and R500 configurations.
All configurations use DINOv3, are initially trained on ProGAN,
and subsequently adapt along the same 14-generator sequence.
EvoKnow and R0 operate without replay, while R500 retains
500 historical samples.
Figure~\ref{fig:forgetting_heatmap} shows that EvoKnow achieves
higher final-stage average accuracy and lower average forgetting
than both Wang configurations.
The trajectories also reveal occasional post-acquisition
improvements, including on Deepfake and BigGAN, consistent
with cross-generator knowledge reuse.
These observations motivate AIR's image-conditioned retrieval
of complementary forensic knowledge across the expanding
expert pool beyond a fixed generator--expert association.

\section{Conclusion}

We presented \textbf{EvoKnow}, a replay-free knowledge-allocation
framework for continual AI-generated image detection. EvoKnow preserves
a shared forensic basis learned from base domains, incrementally adds
isolated residual experts for complementary generator-relevant evidence
from limited current-stage generated images, and uses AIR for analytical
expert retrieval through accumulated sufficient statistics. This design
expands detection coverage without retaining or replaying historical
generated images or their instance-level features. Experiments demonstrate
strong cross-generator and in-the-wild generalization, effective few-shot
expansion, and robust continual adaptation. On the 14-stage benchmark,
EvoKnow achieves the highest reported mean stage-wise accuracy and lowest
reported average forgetting among non-Joint methods. These results suggest
that continual detectors can preserve transferable forensic evidence,
incrementally acquire complementary knowledge, and retrieve relevant
expertise without image replay.

\bibliographystyle{unsrtnat}
\bibliography{Main}  






\appendix

\section{Additional Methodological Details}
\label{app:method}

This appendix details the auxiliary losses, implementation protocol, GPB,
LFB, and analytical router updates used by EvoKnow.

\subsection{Detailed Definitions of Auxiliary Multi-view Losses}
\label{app:loss}

This section details the auxiliary losses in
Eq.~\ref{eq:objective}, which are shared by Phases~I and II. Let
$\mathcal{B}$ denote a mini-batch and
$\mathcal{B}_{+}\subseteq\mathcal{B}$ its generated-image subset.
Expectations are averaged over the indicated samples.

\paragraph{Patch-response Uniformity.}
For a generated image $x$ and a view
$v\in\{g,l_1,\ldots,l_K\}$, let $s^v(x)\in\mathbb{R}^{N}$ denote its
auxiliary patch-logit vector, where $N$ is the number of patches. We
obtain a spatial response distribution
\begin{equation}
a^v(x)=\operatorname{softmax}\left(\frac{s^v(x)}{\tau}\right),
\end{equation}
where $\tau$ is a temperature parameter, set to $1$ in all experiments.
Let $u\in\mathbb{R}^{N}$ be the uniform distribution, with $u_j=1/N$.
The uniformity loss is
\begin{equation}
\mathcal{L}_{\mathrm{uni}}
=
\mathbb{E}_{x\sim\mathcal{B}_{+}}
\left[
D_{\mathrm{KL}}\big(a^g(x)\|u\big)
+
\frac{1}{K}\sum_{k=1}^{K}
D_{\mathrm{KL}}\big(a^{l_k}(x)\|u\big)
\right],
\label{eq:uniformity}
\end{equation}
where
\begin{equation}
D_{\mathrm{KL}}(a\|u)
=
\sum_{j=1}^{N}a_j\log\frac{a_j}{u_j}.
\end{equation}
This loss is applied only to generated images and is set to zero when
$\mathcal{B}_{+}$ is empty.

\paragraph{Weak Patch-level Supervision.}
To ensure that spatially distributed patch responses remain
discriminative, we weakly supervise them using image-level real/fake
labels. For each image $x$, we average the patch logits within the
global view and across all local views:
\begin{equation}
\bar{s}_g(x)
=
\frac{1}{N}\sum_{j=1}^{N}s_j^g(x),
\qquad
\bar{s}_l(x)
=
\frac{1}{KN}\sum_{k=1}^{K}\sum_{j=1}^{N}s_j^{l_k}(x).
\end{equation}
The weak patch-level classification loss is
\begin{equation}
\mathcal{L}_{\mathrm{patch}}
=
\mathbb{E}_{(x,y)\sim\mathcal{B}}
\left[
\ell_{\mathrm{BCE}}\big(\bar{s}_g(x),y\big)
+
\ell_{\mathrm{BCE}}\big(\bar{s}_l(x),y\big)
\right],
\label{eq:patch_supervision}
\end{equation}
where
\begin{equation}
\ell_{\mathrm{BCE}}(s,y)
=
-y\log\sigma(s)
-
(1-y)\log\big(1-\sigma(s)\big).
\end{equation}
This objective requires only image-level annotations and does not use
patch-level labels.

\paragraph{Cross-view Consistency.}
The global and local views describe the authenticity of the same image
and should therefore yield consistent real-versus-generated decisions.
Let $o_g(x)$, $o_l(x)$, and $o_{l_k}(x)$ denote the two-class logits of
the global view, aggregated-local representation, and $k$-th local
view, respectively. We define the generated-versus-real margin as
\begin{equation}
m_v(x)=o_{v,1}(x)-o_{v,0}(x).
\end{equation}
The cross-view consistency loss is
\begin{equation}
\mathcal{L}_{\mathrm{cons}}
=
\mathbb{E}_{(x,y)\sim\mathcal{B}}
\left[
\frac{1}{K}\sum_{k=1}^{K}
\big(m_{l_k}(x)-m_g(x)\big)^2
+
\big(m_l(x)-m_g(x)\big)^2
\right].
\label{eq:consistency}
\end{equation}
The first term aligns individual local views with the global view,
whereas the second aligns the aggregated-local representation with the
global prediction. Together with image-level classification
supervision, these objectives encourage spatially distributed,
discriminative, and cross-view-consistent forensic responses.

\subsection{Implementation and No-Historical-Generated-Image-Replay Protocol}
\label{sec:implementation-details}

This section specifies the architecture, optimization, and data-access
protocol. Unless stated otherwise, experiments use the settings in
Table~\ref{tab:implementation_details}.

\paragraph{Model Configuration.}
We use DINOv3-L/16 with $224\times224$ inputs, one global view, and four
local views. The first six encoder blocks form the frozen shared forensic
learner and define the routing cut, whose patch-token output is denoted by
$S(x)$. AIR derives the $10{,}000$-dimensional routing feature
$\mathbf{z}(x)$ by mean-pooling $S(x)$, followed by a fixed Gaussian
projection, ReLU, and $\ell_2$ normalization. Transformer blocks after
the routing cut belong to the base or incremental residual experts.
Each incremental residual expert uses rank-$8$ LoRA modules, with GPB and
LFB inserted into three deeper blocks.

\paragraph{Three-phase Optimization and Data Access.}
In Phase~I, shared LoRA modules and the base expert $U_0$ are trained on
the base domain, whose generated images also initialize AIR's expert-$0$
target. In Phase~II, only the newly added incremental residual expert is
optimized; the shared forensic learner and all prior experts remain
frozen. Unless otherwise specified, each arriving generator provides ten
generated images.

At each stage, we sample $1{,}000$ task-agnostic real images from a
predefined training pool for negative-class supervision. The pool is
disjoint from evaluation splits, independent of generator domains, and
excluded from AIR fitting. In Phase~III, AIR is updated using only
current-stage generated images and the new expert identifier. Historical
generated images and instance-level routing features are discarded after
each stage; only aggregated router statistics and learned parameters are
retained.

In Phase III, AIR is expanded using only current-stage generated images
and the identifier of the newly added incremental residual expert. After each stage,
historical generated images and all instance-level routing features are
discarded. Routing information from earlier stages is preserved only
through aggregated sufficient statistics and learned router parameters,
as detailed in Appendix~\ref{sec:air-details} and
Appendix~\ref{sec:router-state}.

\begin{table}[t]
\centering
\small
\caption{Implementation details and hyperparameter settings.}
\label{tab:implementation_details}
\setlength{\tabcolsep}{4pt}
\renewcommand{\arraystretch}{1.08}
\begin{tabular}{llc}
\toprule
Phase & Hyperparameter & Value \\
\midrule
\multirow{8}{*}{Shared knowledge learning}
& Backbone & DINOv3-L/16 \\
& Input resolution & $224\times224$ \\
& Global / local views & $1$ / $4$ \\
& Optimizer & AdamW \\
& Learning rate & $5\times10^{-6}$ \\
& Weight decay & $10^{-2}$ \\
& Batch size & $8$ \\
& Optimization steps & $3{,}000$ \\
\midrule
\multirow{8}{*}{Residual knowledge learning}
& Generated images per stage & $10$ \\
& Task-agnostic real images sampled per stage & $1{,}000$ \\
& Optimizer & AdamW \\
& Learning rate & $5\times10^{-6}$ \\
& Batch size & $4$ \\
& Training epochs & $2$ \\
& LoRA rank & $8$ \\
& Trainable modules & LoRA, GPB, LFB, and heads \\
\midrule
\multirow{6}{*}{Analytical router}
& Feature source & First $6$ encoder blocks \\
& Routing feature dimension & $10{,}000$ \\
& Feature mapping & Gaussian projection + ReLU + $\ell_2$ norm. \\
& Update rule & Recursive least squares \\
& Ridge coefficient $\lambda_{\mathrm{RLS}}$ & $1{,}000$ \\
& Router fitting data & Current generated images only \\
\midrule
\multirow{4}{*}{Multi-view objective}
& $\lambda_{\mathrm{cons}}$ & $0.2$ \\
& $\lambda_{\mathrm{uni}}$ & $0.02$ \\
& $\lambda_{\mathrm{patch}}$ & $0.1$ \\
& $(\omega_g,\omega_l,\omega_f)$ & $(0.3,0.3,1.0)$ \\
\bottomrule
\end{tabular}
\end{table}

\subsection{Resource Costs and Routing Dimension}
\label{sec:supp-resource-costs}

\noindent\textbf{Resource Scaling.}
We measure resource costs along the 14-stage continual-learning
sequence using DINOv3-L/16 on a single NVIDIA L20 GPU, with
routing dimension $m=10{,}000$.
Table~\ref{tab:supp-resource-scaling} reports checkpoint storage,
incremental training time, AIR initialization or update time,
and end-to-end inference latency.
From one to fourteen tasks, saved model size increases from
1280.16 to 1527.02 MB, while inference latency increases from
153.98 to 162.36 ms/image.
New-expert training takes approximately 86 seconds at the
measured incremental stages, and mean AIR update time remains
below one second.
Only the selected expert is executed for each image, although
routing computation grows with the expert pool.

\begin{table*}[t]
\centering
\scriptsize
\setlength{\tabcolsep}{6pt}
\renewcommand{\arraystretch}{1.20}
\caption{
Resource costs at selected stages of the 14-stage
continual-learning sequence with DINOv3-L/16 and routing
dimension $m=10{,}000$.
Saved model size includes the backbone, all experts, and AIR
state. Expert training time refers to the newly added expert
at the indicated stage.
Inference latency is measured at batch size 1 on a single
NVIDIA L20 GPU with identical numerical precision across settings.
Timing results are reported as mean $\pm$ standard deviation.
}
\label{tab:supp-resource-scaling}
\resizebox{\textwidth}{!}{%
\begin{tabular}{@{}crrrr@{}}
\toprule
\shortstack{Observed\\tasks}
& \shortstack{Saved model\\size (MB)}
& \shortstack{Incremental expert\\training (s)}
& \shortstack{AIR initialization\\or update (s)}
& \shortstack{End-to-end inference\\(ms/image)} \\
\midrule
1 (base)
& 1280.16
& --
& $0.32 \pm 0.04$
& $153.98 \pm 0.18$ \\

2
& 1337.00
& $85.93 \pm 0.90$
& $0.77 \pm 0.74$
& $160.47 \pm 0.18$ \\

8
& 1432.01
& $85.64 \pm 0.77$
& $0.82 \pm 0.81$
& $159.40 \pm 1.76$ \\

14
& 1527.02
& $85.80 \pm 0.63$
& $0.58 \pm 0.03$
& $162.36 \pm 0.36$ \\
\bottomrule
\end{tabular}%
}
\end{table*}

\subsection{Global Pattern-aware Block: Differentiable Structural Comparison}
\label{sec:supp-gpb}

The Global Pattern-aware Block (GPB) augments global-view features with
explicit structural evidence by comparing each patch with its spatial
neighbors. We use a smooth relaxation of hard comparisons so that the
detection objective can optimize structural responses end-to-end.

Let $\mathbf{F}\in\mathbb{R}^{HW\times C}$ denote the patch-token
features after a Transformer block, excluding class and storage tokens.
An affine down-projection $D_g$, followed by LayerNorm and spatial
reshaping, produces
\begin{equation}
\mathbf{Z}_g
=
\operatorname{reshape}\!\left(
\operatorname{LN}(D_g(\mathbf{F}))
\right),
\end{equation}
where $\mathbf{Z}_g$ has $C'=\lfloor C/8\rfloor$ channels. For a patch
location $i$ and one of its eight neighbors $j\in\mathcal{N}_8(i)$,
let $d_{ij}=Z_{g,j}-Z_{g,i}$ denote their feature difference in a given
channel. A hard comparison would produce
$h(d)=\mathbb{1}[d\geq0]$, whose derivative is zero away from the
threshold and undefined at $d=0$. GPB therefore replaces the hard
comparison with the following differentiable relaxation:

\begin{equation}
\begin{aligned}
h(d)=\mathbb{1}[d\geq0]
&\quad\longrightarrow\quad
s_k(d)=\sigma(kd),\\
k&=\operatorname{clip}(e^\eta,1,50),
\end{aligned}
\label{eq:supp-gpb-comparison}
\end{equation}
where $\sigma$ is the sigmoid function and $\eta$ is a learnable
scalar initialized to $\log 10$.
The corresponding feature-difference derivative is
\begin{equation}
\frac{\partial s_k(d)}{\partial d}
=k\,s_k(d)\bigl(1-s_k(d)\bigr).
\label{eq:supp-gpb-gradient}
\end{equation}
This derivative allows the detection objective to adjust both the
center and neighboring features, with opposite signs through their
difference.
Larger $k$ sharpens the transition and concentrates sensitivity near
$d=0$, whereas smaller $k$ spreads sensitivity over a wider range
of feature differences.
The derivative at $d=0$ is $k/4$.
The sharpness is trainable within its clipping bounds; clipping can
suppress its gradient outside that interval.

\begin{figure*}[!t]
\centering
\includegraphics[width=\textwidth]{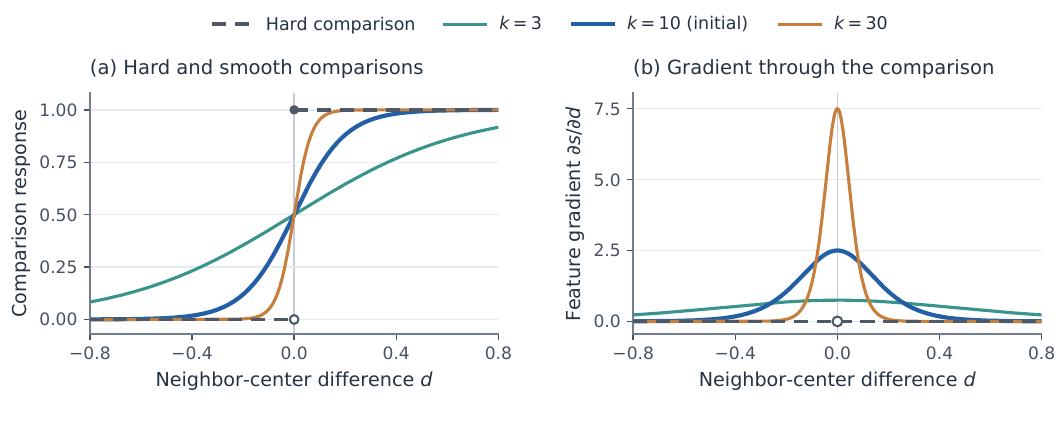}
\caption{
Analytical illustration of the structural comparison in GPB.
(a) A hard neighbor--center comparison and its sigmoid relaxations.
(b) Derivatives with respect to the feature difference.
The initial sharpness is $k=10$; the other values illustrate how
sharpness changes the transition and gradient concentration.
The hard comparison has zero derivative for $d\neq0$ and no ordinary
derivative at $d=0$, indicated by an open marker.
}
\label{fig:supp-gpb-relaxation}
\end{figure*}

GPB applies the comparison independently to each bottleneck channel
and averages the eight neighboring responses:
\begin{equation}
\mathbf{R}_{g,i}
=\frac{1}{8}\sum_{j\in\mathcal{N}_8(i)}
\sigma\!\left(k(\mathbf{Z}_{g,j}-\mathbf{Z}_{g,i})\right).
\label{eq:supp-gpb-response}
\end{equation}
Reflection padding defines the neighborhood at feature-map boundaries.
The averaging produces a local relative-order response inspired by
binary texture comparisons; it does not encode the neighbors as a
directional, bit-weighted LBP integer.
The response is combined with the bottleneck features and projected
back to the backbone dimension:
\begin{equation}
\begin{aligned}
\Delta\mathbf{F}^{\mathrm{GPB}}
&=\gamma_g\,
\operatorname{LN}\!\left(
U_g\!\left(\operatorname{flat}(\mathbf{Z}_g+\mathbf{R}_g)\right)
\right),\\
\mathbf{F}'&=\mathbf{F}+\Delta\mathbf{F}^{\mathrm{GPB}},
\end{aligned}
\label{eq:supp-gpb-residual}
\end{equation}
where $U_g$ is an affine up-projection and $\operatorname{flat}$
restores the patch-token sequence.
The learnable residual scale $\gamma_g$ is initialized to $10^{-3}$
in the block wrapper.
GPB directly modifies only patch tokens; its residual entries for
the class and storage tokens are zero.
Subsequent attention layers can still propagate the updated patch
information to those tokens.
Figure~\ref{fig:supp-gpb-relaxation} illustrates the comparison
responses and their gradients.

\subsection{Local Frequency-aware Block: Learnable Spectral Selection}
\label{sec:supp-lfb}

The Local Frequency-aware Block (LFB) augments local-view features with
learnable spectral residuals. It uses a smooth high-pass transition so
that the cutoff can be optimized jointly with the detection objective.

An independent affine down-projection $D_f$ and LayerNorm produce
the bottleneck map
$\mathbf{Z}_f=\operatorname{reshape}(\operatorname{LN}(D_f(\mathbf{F})))$,
again with $C'=\lfloor C/8\rfloor$ channels.
For a Fourier bin with frequencies $(f_x,f_y)$, define
$\rho=\sqrt{f_x^2+f_y^2}$.
The implementation uses the frequencies returned by
\texttt{rfftfreq} and \texttt{fftfreq}, measured in cycles per patch
spacing; $\rho$ is not rescaled to $[0,1]$.
A hard high-pass mask is
$M_{\mathrm{hard}}(\rho;c)=\mathbb{1}[\rho\geq c]$.
At a fixed Fourier bin, this mask has zero derivative with respect
to $c$ away from the threshold.
Although filtering with a fixed mask remains differentiable with
respect to its input, the hard cutoff does not provide an ordinary
gradient for learning its location.

\begin{figure*}[!t]
\centering
\includegraphics[width=\textwidth]{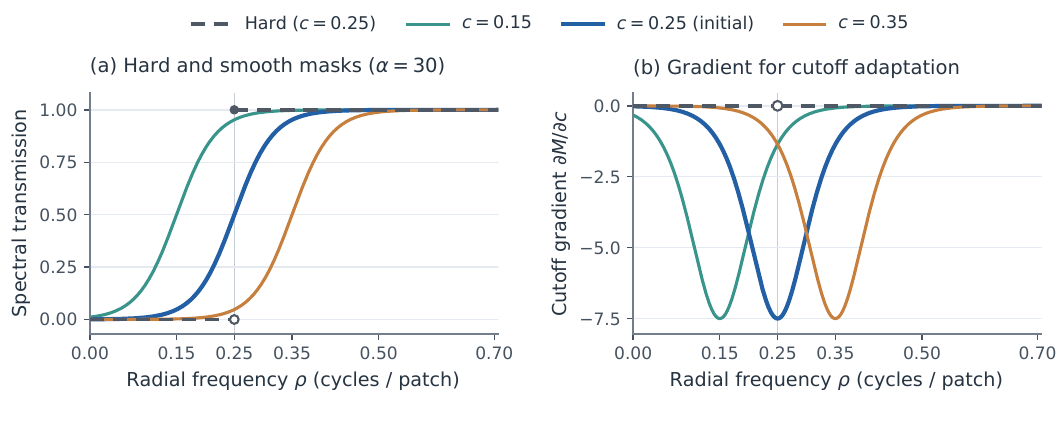}
\caption{
Analytical illustration of spectral selection in LFB.
(a) A hard high-pass mask at $c=0.25$ and smooth masks with
$\alpha=30$ at three cutoff values.
(b) Their derivatives with respect to the cutoff.
The initial cutoff is $c=0.25$.
The smooth mask provides cutoff gradients near the transition;
the hard mask has zero cutoff derivative away from $\rho=c$.
Frequency is measured on the patch-feature grid.
}
\label{fig:supp-lfb-relaxation}
\end{figure*}

LFB therefore uses a smooth radial mask:
\begin{equation}
\begin{aligned}
M_{\mathrm{hard}}(\rho;c)=\mathbb{1}[\rho\geq c]
&\quad\longrightarrow\quad
M(\rho;c)=\sigma\!\left(\alpha(\rho-c)\right),\\
c&=\tfrac{1}{2}\sigma(\theta),
\end{aligned}
\label{eq:supp-lfb-mask}
\end{equation}
where $\theta$ is learnable and $\alpha=30$ is fixed.
The parameterization constrains $c$ to $(0,0.5)$, with the initial
value $c=0.25$ obtained from $\theta=0$.
The same scalar cutoff and radial mask are shared across channels
within each LFB.
The derivatives are
\begin{equation}
\begin{aligned}
\frac{\partial M}{\partial c}
&=-\alpha M(1-M),\\
\frac{\partial M}{\partial\theta}
&=-\frac{\alpha}{2}\,
\sigma(\theta)\bigl(1-\sigma(\theta)\bigr)\,M(1-M).
\end{aligned}
\label{eq:supp-lfb-gradient}
\end{equation}
These derivatives provide a gradient path from the detection loss to
the cutoff parameter, with the strongest mask sensitivity near
$\rho=c$.
The smooth transition also retains partial responses near the cutoff.
Figure~\ref{fig:supp-lfb-relaxation} compares hard and smooth masks
and illustrates their cutoff sensitivities.

\paragraph{Frequency residual.}
LFB applies the mask to the real-valued two-dimensional Fourier
transform of each bottleneck feature channel:
\begin{equation}
\begin{aligned}
\mathbf{Z}_{\mathrm{hp}}
&=\mathcal{F}^{-1}_{r}
\!\left(
M\odot\mathcal{F}_{r}(\mathbf{Z}_f)
\right),\\
\Delta\mathbf{F}^{\mathrm{LFB}}
&=\gamma_f U_f\!\left(\operatorname{flat}(\mathbf{Z}_{\mathrm{hp}})\right).
\end{aligned}
\label{eq:supp-lfb-residual}
\end{equation}
Here, $\mathcal{F}_{r}$ and $\mathcal{F}^{-1}_{r}$ correspond to
\texttt{rfft2} and \texttt{irfft2} with orthonormal normalization;
the inverse transform uses the original spatial dimensions.
The transforms are evaluated in FP32 and the filtered features
are cast back to the input dtype before the up-projection.
The residual is added to the patch-token output of the wrapped
Transformer block, with zero residual entries for the class and
storage tokens.
The wrapper initializes the learnable scale $\gamma_f$ to $10^{-3}$.
This attenuates the initial residual without making it identically zero.

\paragraph{Placement and branch selection.}
For DINOv3-L/16, the two side modules are attached after blocks
$7$, $13$, and $19$ in one-based indexing, corresponding to
indices $6$, $12$, and $18$ in the code.
Their bottleneck width is $1024/8=128$.
In the global/local design, GPB processes the global branch and
LFB processes the local branches.
The wrapper exposes independent activation flags for these modules;
when both are enabled in one call, the frequency module processes
the tokens after the structural residual has been added.
The smooth operators above remain part of the feature computation;
their hard counterparts serve as reference operations for explaining
the relaxation.

\subsection{Analytical Incremental Router Details}
\label{sec:air-details}

This section supplements the AIR overview in Section~\ref{sec:router}
by specifying its base-expert initialization and recursive expansion.
AIR selects among the base expert $U_0$ and incremental residual experts
$\{U_i\}_{i=1}^{t}$ in a fixed routing-feature space.

Given the routing-cut representation $S(x)$ produced by the frozen shared
forensic learner, AIR computes $\mathbf{z}(x)=g(S(x))$, where
$g(\cdot)$ mean-pools the patch tokens of $S(x)$ and applies a fixed
Gaussian projection, ReLU activation, and $\ell_2$ normalization. Since
both the shared forensic learner and $g(\cdot)$ remain fixed after base
training, routing features are comparable across incremental stages.

\paragraph{Base-router Initialization.}
AIR is initialized using generated images from the base domain, all
assigned to the base expert $U_0$. Let
$\mathbf{Z}_0\in\mathbb{R}^{n_0\times m}$ contain their routing
features, where $m$ is the routing-feature dimension, and let
$\mathbf{Y}_0\in\mathbb{R}^{n_0\times1}$ be the corresponding
single-column target matrix. We initialize
\begin{equation}
\begin{aligned}
\mathbf{G}_0
&=
\lambda_{\mathrm{RLS}}\mathbf{I}
+
\mathbf{Z}_0^\top\mathbf{Z}_0,\\
\mathbf{R}_0
&=
\mathbf{G}_0^{-1},\\
\mathbf{Q}_0
&=
\mathbf{Z}_0^\top\mathbf{Y}_0,\\
\mathbf{W}_0
&=
\mathbf{R}_0\mathbf{Q}_0,
\end{aligned}
\label{eq:air_init}
\end{equation}
where $\lambda_{\mathrm{RLS}}$ is the ridge regularization coefficient.
The recursive implementation retains $\mathbf{R}_0$ rather than
explicitly recomputing the inverse of $\mathbf{G}_0$ at later stages.

\paragraph{Incremental Router Expansion.}
At stage $t$, let
$\mathbf{Z}_t\in\mathbb{R}^{n_t\times m}$ contain routing features of
the current-stage generated images. Their one-hot expert-selection
targets $\mathbf{Y}_t\in\mathbb{R}^{n_t\times(t+1)}$ assign all current
samples to the newly added expert $U_t$, while retaining columns for the base expert and previously added
incremental residual experts. These targets
supervise expert retrieval rather than exact generator-source
attribution. To expand the output space, we append a zero column to the
previous feature--target statistic:
\begin{equation}
\widetilde{\mathbf{Q}}_{t-1}
=
\left[
\mathbf{Q}_{t-1},\mathbf{0}
\right].
\end{equation}
AIR then updates its sufficient statistics and router state as
\begin{equation}
\begin{aligned}
\mathbf{G}_t
&=
\mathbf{G}_{t-1}
+
\mathbf{Z}_t^\top\mathbf{Z}_t,\\
\mathbf{R}_t
&=
\mathbf{R}_{t-1}
-
\mathbf{R}_{t-1}\mathbf{Z}_t^\top
\left(
\mathbf{I}
+
\mathbf{Z}_t\mathbf{R}_{t-1}\mathbf{Z}_t^\top
\right)^{-1}
\mathbf{Z}_t\mathbf{R}_{t-1},\\
\mathbf{Q}_t
&=
\widetilde{\mathbf{Q}}_{t-1}
+
\mathbf{Z}_t^\top\mathbf{Y}_t,\\
\mathbf{W}_t
&=
\mathbf{R}_t\mathbf{Q}_t,
\end{aligned}
\label{eq:router_update}
\end{equation}
where $\mathbf{R}_t=\mathbf{G}_t^{-1}$ and
$\mathbf{W}_t\in\mathbb{R}^{m\times(t+1)}$. The update of
$\mathbf{R}_t$ follows the Woodbury identity and avoids explicitly
inverting the $m\times m$ Gram matrix at every stage. With fixed routing
features, a fixed ridge coefficient, and the same one-hot routing
targets, this recursive update yields the same solution as refitting the
AIR ridge-regression objective on all previously observed routing
samples. This equivalence applies only to the AIR routing objective; it
does not involve joint optimization of expert parameters. The update
avoids storing historical routing samples and does not require
gradient-based router optimization.

\paragraph{Routing and Detection.}
AIR is fitted only on generated images because it retrieves the expert
whose residual forensic evidence is most compatible with the input.
Its objective is expert retrieval for binary detection rather than exact
generator-source attribution. Real images
are excluded from router fitting but supervise every expert for binary
real/fake detection. At inference, AIR selects
\begin{equation}
r_t(x)
=
\arg\max_{0\leq i\leq t}
\mathbf{z}(x)^\top\mathbf{w}_{t,i},
\end{equation}
where $\mathbf{w}_{t,i}$ is the $i$-th column of
$\mathbf{W}_t$. The selected expert outputs
\begin{equation}
p_t(y=1\mid x)
=
U_{r_t(x)}(S(x)).
\end{equation}

\section{Supplementary Experiments}
\label{sec:supplementary}

\begin{table*}[h]
\centering
\small
\caption{
Complete generator-wise detection trajectory on GenImage.
Generators arrive in the column order.
Seen averages accuracy over the generators observed by the current stage;
All averages accuracy over all eight generators, including those not yet
observed.
Bold marks the highest values in the Seen and All columns.
All entries are percentages; averages are computed from the displayed
generator-wise results.
}
\label{tab:supp-trajectory}
\resizebox{\textwidth}{!}{%
\begin{tabular}{lcccccccccc}
\toprule
Stage & SD v1.4 & BigGAN & VQDM & SD v1.5 & Wukong
& ADM & GLIDE & Midjourney & Seen & All \\
\midrule
T1 & 99.69 & 93.78 & 95.67 & 99.69 & 99.74
& 78.05 & 95.52 & 93.10 & \textbf{99.69} & 94.41 \\
T2 & 100.00 & 97.54 & 97.78 & 99.93 & 99.96
& 83.19 & 97.59 & 94.43 & 98.77 & 96.30 \\
T3 & 100.00 & 97.54 & 97.77 & 99.93 & 99.96
& 83.19 & 97.59 & 94.42 & 98.44 & 96.30 \\
T4 & 100.00 & 97.54 & 97.52 & 99.93 & 99.96
& 82.87 & 97.59 & 94.31 & 98.75 & 96.22 \\
T5 & 100.00 & 97.32 & 97.42 & 99.93 & 99.96
& 81.54 & 97.03 & 94.08 & 98.93 & 95.91 \\
T6 & 100.00 & 97.41 & 97.54 & 99.93 & 99.96
& 83.96 & 97.47 & 94.72 & 96.47 & \textbf{96.37} \\
T7 & 100.00 & 97.18 & 97.46 & 99.93 & 99.96
& 83.83 & 97.34 & 94.64 & 96.53 & 96.29 \\
T8 & 100.00 & 97.18 & 97.46 & 99.93 & 99.96
& 83.85 & 97.35 & 94.78 & 96.31 & 96.31 \\
\bottomrule
\end{tabular}%
}
\end{table*}

This section provides supplementary experiments supporting EvoKnow's
continual knowledge-evolution mechanism. We first analyze complete
continual trajectories and historical knowledge retention, then study
the effect of the adaptation budget on cross-generator transfer.
Finally, we examine the persistent state, storage scaling, and
auxiliary attribution behavior of the analytical router.

\subsection{Continual Learning Trajectories and Knowledge Retention}
\label{sec:task-order}

\begin{table*}[!t]
\centering
\small
\setlength{\tabcolsep}{4pt}
\caption{
Stage-wise continual detection performance on GenImage.
Each generator denotes its arrival stage.
AA averages accuracy over the generators observed so far;
AF measures the average signed drop from each previously learned generator’s best post-acquisition accuracy before the current stage to its accuracy at the current stage.
AA is reported in percentages and AF in percentage points.
AF is undefined at the first stage.
Final-stage results are bolded.
}
\label{tab:supp-continual-metrics}
\resizebox{\textwidth}{!}{%
\begin{tabular}{l*{16}{c}}
\toprule
Method
& \multicolumn{2}{c}{SD v1.4}
& \multicolumn{2}{c}{BigGAN}
& \multicolumn{2}{c}{VQDM}
& \multicolumn{2}{c}{SD v1.5}
& \multicolumn{2}{c}{Wukong}
& \multicolumn{2}{c}{ADM}
& \multicolumn{2}{c}{GLIDE}
& \multicolumn{2}{c}{Midjourney} \\
\cmidrule(lr){2-3}
\cmidrule(lr){4-5}
\cmidrule(lr){6-7}
\cmidrule(lr){8-9}
\cmidrule(lr){10-11}
\cmidrule(lr){12-13}
\cmidrule(lr){14-15}
\cmidrule(lr){16-17}
& AA $\uparrow$ & AF $\downarrow$
& AA $\uparrow$ & AF $\downarrow$
& AA $\uparrow$ & AF $\downarrow$
& AA $\uparrow$ & AF $\downarrow$
& AA $\uparrow$ & AF $\downarrow$
& AA $\uparrow$ & AF $\downarrow$
& AA $\uparrow$ & AF $\downarrow$
& AA $\uparrow$ & AF $\downarrow$ \\
\midrule
\rowcolor{blue!5}
EvoKnow
& 99.69 & --
& 98.77 & $-0.31$
& 98.44 & 0.00
& 98.75 & 0.08
& 98.93 & 0.14
& 96.47 & 0.07
& 96.53 & 0.13
& \textbf{96.31} & \textbf{0.11} \\
\bottomrule
\end{tabular}%
}
\end{table*}

\begin{table*}[t]
\centering
\begin{minipage}[t]{0.49\textwidth}
\vspace{0pt}
\centering
\small
\caption{
Generator-wise retention at the final stage on GenImage.
Historical best accuracy is the highest accuracy attained after
the corresponding generator is learned and before the final
stage, T8. Drop is the historical best accuracy minus the final
accuracy, measured in percentage points; a negative value
indicates improvement beyond the previous best.
AF averages these drops over the seven previously learned
generators, excluding Midjourney, which arrives at T8.
}
\label{tab:supp-forgetting}
\resizebox{\linewidth}{!}{%
\begin{tabular}{lccc}
\toprule
Generator
& \shortstack{Historical best\\accuracy (\%)}
& \shortstack{Final\\accuracy (\%)}
& \shortstack{Drop\\(pp)} \\
\midrule
SD v1.4 & 100.00 & 100.00 & 0.00 \\
BigGAN  & 97.54  & 97.18  & 0.36 \\
VQDM    & 97.77  & 97.46  & 0.31 \\
SD v1.5 & 99.93  & 99.93  & 0.00 \\
Wukong  & 99.96  & 99.96  & 0.00 \\
ADM     & 83.96  & 83.85  & 0.11 \\
GLIDE   & 97.34  & 97.35  & $-0.01$ \\
\midrule
AF $\downarrow$ & -- & -- & \textbf{0.11} \\
\bottomrule
\end{tabular}%
}
\end{minipage}\hfill
\begin{minipage}[t]{0.49\textwidth}
\vspace{0pt}
\centering
\small
\caption{
Persistent state of the analytical router and incremental residual
knowledge units.
$m$ is the routing-feature dimension, $d$ is the input dimension
of the fixed projection, $M$ is the number of residual knowledge
units, and $p$ is the number of additional parameters per unit.
The frozen shared backbone is excluded.
}
\label{tab:supp-router-state}
\resizebox{\linewidth}{!}{%
\begin{tabular}{lcc}
\toprule
State & Shape / count & Storage order \\
\midrule
$\mathbf{R}$ (Gram inverse)
& $m\times m$ & $O(m^2)$ \\

$\mathbf{Q}$ (feature--target statistic)
& $m\times M$ & $O(mM)$ \\

$\mathbf{W}=\mathbf{R}\mathbf{Q}$ (router weights)
& $m\times M$ & $O(mM)$ \\

$\mathbf{P}$ (fixed projection)
& $m\times d$ & $O(md)$ \\

\shortstack[l]{Incremental residual\\expert parameters}
& $M$ units & $O(Mp)$ \\

Unit metadata
& $M$ entries & $O(M)$ \\
\bottomrule
\end{tabular}%
}
\end{minipage}
\end{table*}

We analyze how EvoKnow evolves as incremental residual experts are added
sequentially. The GenImage sequence starts from the Phase-I base domain, SD v1.4, followed by BigGAN, VQDM, SD v1.5, Wukong, ADM, GLIDE, and Midjourney.
The shared forensic learner is initialized on SD v1.4 and remains frozen
throughout subsequent adaptation. The generator arrival order in this supplementary experiment differs from that used for the main results in Table 1; Tables 5–7 report the trajectory and retention results under this alternative order.

After each stage, we evaluate the detector on all eight generator
domains, including generators not yet observed at that stage. Let
$B_{t,j}$ denote detection accuracy on generator $j$ after stage $t$.
We distinguish accuracy on observed domains and all evaluation domains:
\begin{equation}
\mathrm{Seen}_t
=
\frac{1}{t}\sum_{j=1}^{t}B_{t,j},
\qquad
\mathrm{All}_t
=
\frac{1}{8}\sum_{j=1}^{8}B_{t,j}.
\label{eq:supp-accuracy}
\end{equation}
Here, $\mathrm{Seen}_t$ measures performance on acquired domains,
whereas $\mathrm{All}_t$ additionally measures transfer to generators
not yet observed. At the final stage, the two quantities coincide.


Table~\ref{tab:supp-trajectory} reports the complete generator-wise
trajectory. The all-domain average improves overall from 94.41\% at the first stage
to 96.31\% after the final stage, despite moderate fluctuations as
domains with different difficulty levels are introduced. A decrease in
$\mathrm{Seen}_t$ does not necessarily indicate forgetting because a
new and potentially more difficult generator is added to the average at
every stage. For example, when ADM arrives at Stage~6,
$\mathrm{Seen}_t$ decreases while $\mathrm{All}_t$ increases and AF
remains low.

Tables~\ref{tab:supp-continual-metrics}
and~\ref{tab:supp-forgetting} summarize knowledge retention.
Here, AF averages the drop from each previously learned
generator's best post-acquisition accuracy before T8 to its
accuracy at T8; negative values indicate improvement beyond
the previous best.
The final AF of 0.11 percentage points indicates limited
forgetting. Since the shared forensic learner and existing
experts remain frozen, historical-domain performance can
change through the evolving router's expert selection.

\begin{table*}[!t]
\centering
\scriptsize
\renewcommand{\arraystretch}{0.93}
\setlength{\tabcolsep}{2.0pt}
\caption{
UniversalFakeDetect detection accuracy (\%) under zero-shot and
few-shot knowledge expansion. Existing-detector results follow their
reported ProGAN-based training protocols. For EvoKnow, the adaptation
budget denotes the number of generated images per arriving generator in
the GenImage stream; UniversalFakeDetect is used only for evaluation.
The knowledge-unit organization and arrival sequence are fixed across
adaptation budgets. The final column reports average accuracy and
population standard deviation across the 14 displayed generators; the
standard deviation measures cross-generator variation rather than
variation across repeated runs.
}
\label{tab:fewshot_full}
\resizebox{\textwidth}{!}{%
\begin{tabular}{llccccccccccccccc}
\toprule
Method & Adaptation budget
& \multicolumn{6}{c}{GAN}
& Guided
& \multicolumn{2}{c}{LDM}
& \multicolumn{4}{c}{GLIDE}
& DALL-E & Ave. $\pm$ Std. \\
\cmidrule(lr){3-8}
\cmidrule(lr){10-11}
\cmidrule(lr){12-15}
& & ProG. & Cycle & BigG. & Style & GauG. & StarG.
& & 200 & 200-cfg
& 100 & 100/27 & 50/27 & 100/10 & & \\
\midrule
CNNSpot & CVPR'20
& 99.99 & 85.20 & 70.20 & 85.70 & 78.95 & 91.70
& 65.66 & 60.07 & 54.03 & 54.96 & 54.14 & 60.78 & 63.80
& 55.58 & 70.05$\pm$14.90 \\
PatchFor. & ECCV'20
& 75.03 & 68.97 & 68.47 & 79.16 & 64.23 & 63.94
& 68.52 & 67.41 & 76.50 & 76.10 & 75.77 & 74.81 & 73.28
& 67.91 & 71.44$\pm$4.73 \\
Co-occur. & EI'20
& 97.70 & 63.15 & 53.75 & 92.50 & 51.10 & 54.70
& 69.90 & 60.50 & 70.70 & 70.55 & 71.00 & 70.25 & 69.60
& 67.55 & 68.78$\pm$12.69 \\
Freq-spec & WIFS'19
& 49.90 & 99.90 & 50.50 & 49.90 & 50.30 & 99.70
& 50.40 & 50.90 & 50.40 & 50.40 & 50.30 & 51.70 & 51.40
& 50.00 & 57.55$\pm$17.26 \\
F3Net & ECCV'20
& 99.38 & 76.38 & 65.33 & 92.56 & 58.10 & 100.00
& 83.05 & 69.20 & 68.15 & 75.35 & 68.80 & 81.65 & 83.25
& 66.30 & 77.68$\pm$12.48 \\
UnivFD & CVPR'23
& 100.00 & 98.50 & 94.50 & 82.00 & 99.50 & 97.00
& 78.14 & 70.03 & 94.19 & 73.76 & 94.36 & 79.07 & 79.85
& 86.78 & 87.69$\pm$9.98 \\
LGrad & CVPR'23
& 99.84 & 85.39 & 82.88 & 94.83 & 72.45 & 99.62
& 89.55 & 77.50 & 94.20 & 95.85 & 94.80 & 87.40 & 90.70
& 88.35 & 89.53$\pm$7.70 \\
FreqNet & AAAI'24
& 97.90 & 95.84 & 90.45 & 97.55 & 90.24 & 93.41
& 88.15 & 86.70 & 84.55 & 99.58 & 65.56 & 85.69 & 97.40
& 59.06 & 88.01$\pm$11.57 \\
NPR & CVPR'24
& 99.84 & 95.00 & 87.55 & 96.23 & 86.57 & 99.75
& 97.35 & 84.55 & 97.65 & 98.00 & 98.20 & 96.25 & 97.15
& 87.15 & 94.37$\pm$5.19 \\
FatFormer & CVPR'24
& 99.89 & 99.32 & 99.50 & 97.15 & 99.41 & 99.75
& 94.20 & 76.00 & 98.60 & 94.90 & 98.65 & 94.35 & 94.65
& 98.75 & 96.08$\pm$5.95 \\
C2P-CLIP & AAAI'25
& 99.98 & 97.31 & 99.12 & 96.44 & 99.17 & 99.60
& 96.10 & 69.10 & 99.25 & 97.25 & 99.30 & 95.25 & 95.25
& 98.55 & 95.83$\pm$7.58 \\

\midrule
\rowcolor{blue!5}
& 0-shot
& 93.75 & 90.46 & 98.03 & 93.73 & 97.99 & 93.83
& 98.50 & 83.60 & 99.27 & 98.50 & 99.30 & 97.97 & 97.97
& 98.53 & 95.82$\pm$4.27 \\

\rowcolor{blue!5}
& 10-shot
& 96.50 & 94.80 & 99.00 & 96.20 & 98.80 & 98.00
& 99.10 & 91.50 & 99.55 & 99.10 & 99.60 & 98.80 & 98.80
& 99.20 & 97.78$\pm$2.22 \\

\rowcolor{blue!5}
\makecell[c]{\textbf{EvoKnow}}
& 50-shot
& 97.80 & 95.37 & 99.10 & 97.35 & 98.53 & 99.75
& 98.55 & 93.32 & 99.05 & 98.00 & 99.30 & 98.00 & 98.75
& 99.20 & 98.01$\pm$1.67 \\

\rowcolor{blue!5}
& 100-shot
& 98.37 & 97.80 & 99.32 & 97.75 & 99.32 & 99.17
& 99.65 & 94.06 & 99.40 & 99.54 & 99.63 & 99.71 & 99.82
& 99.65 & 98.79$\pm$1.47 \\

\rowcolor{blue!5}
& 300-shot
& 98.54 & 98.50 & 99.50 & 97.55 & 98.75 & 99.55
& 99.85 & 94.47 & 99.58 & 99.27 & 99.54 & 99.17 & 99.20
& 99.84 & 98.81$\pm$1.35 \\
\bottomrule
\end{tabular}%
}
\end{table*}

\begin{table}[!t]
\centering
\scriptsize
\setlength{\tabcolsep}{2.2pt}
\caption{
Effect of incremental residual knowledge organization on GenImage (\%).
Generator-based organization assigns one incremental residual expert per
generator, whereas category-based organization groups generators into
GAN, diffusion, and unknown-source categories. Both settings use ten
generated images per arriving generator. Unlike
Table~\ref{tab:generalization}, \emph{Ave.} here includes all eight
GenImage domains, including the SD v1.4 base domain.
}
\label{tab:unit_construction}
\resizebox{\linewidth}{!}{%
\begin{tabular}{lccccccccc}
\toprule
Organization
& MidJ. & SD1.4 & SD1.5 & ADM & GLIDE & Wuk. & VQDM & BigGAN & Ave. \\
\midrule
Generator-based
& 96.06 & 98.18 & 98.53 & 96.21 & 96.27
& \textbf{98.64} & 97.41 & 93.77 & 96.88 \\

Category-based
& \textbf{97.39} & \textbf{99.32} & \textbf{99.62}
& \textbf{97.30} & \textbf{98.32}
& 98.61 & \textbf{99.21} & \textbf{99.64}
& \textbf{98.68} \\
\midrule
\rowcolor{orange!10}
$\Delta$
& +1.33 & +1.14 & +1.09 & +1.09 & +2.05
& -0.03 & +1.80 & +5.87 & +1.79 \\
\bottomrule
\end{tabular}%
}
\end{table}

\subsection{Few-shot Adaptation and Cross-Generator Transfer}
\label{sec:supp-fewshot}

Starting from the same GenImage-trained shared basis, we vary the number
of generated images available for each incoming generator over
$\{10,50,100,300\}$ while fixing the knowledge-unit organization and
arrival sequence. The resulting detectors are evaluated on
UniversalFakeDetect. Thus, the shot budget refers to the GenImage
adaptation stream, whereas UniversalFakeDetect is used only for
evaluation.

\subsection{Incremental Residual Knowledge Organization}
\label{sec:supp-organization}

Table~\ref{tab:fewshot_full} reports complete generator-wise results.
Average accuracy increases from 95.82\% without adaptation to 97.78\%
with ten-shot adaptation and 98.81\% with 300-shot adaptation.
Meanwhile, across-generator variation decreases substantially, from
4.27 to 1.35 percentage points. The largest gains occur on initially
challenging domains, indicating that additional residual knowledge
primarily compensates for generator-relevant gaps in the shared forensic basis.

The default EvoKnow setting assigns one incremental residual expert to
each arriving generator domain. As an additional analysis, we compare
this generator-based organization with a category-based organization that
groups generators into GAN, diffusion, and unknown-source categories.
Both settings use ten generated images per arriving generator.

Table~\ref{tab:unit_construction} shows that category-based organization
improves seven of eight GenImage domains and raises the all-domain
average by 1.79 percentage points. This result suggests that
incrementally acquired residual knowledge is not entirely
generator-exclusive: related generators can benefit from partially shared
complementary forensic evidence.

\subsection{Analytical Router State, Resource Scaling, and Attribution Diagnostics}
\label{sec:router-state}


\begin{table*}[!t]
\centering
\small
\caption{
Auxiliary source-attribution diagnostic (rank-1 accuracy, \%).
The evaluation includes one real-image class and eight generator
classes. The real label is determined by the binary detection head,
whereas generator labels are supplied by the analytical router.
Avg.\ is the unweighted mean over all nine classes. Big, Mid, Wuk, GLI,
and VQ abbreviate BigGAN, Midjourney, Wukong, GLIDE, and VQDM,
respectively.
}
\label{tab:supp-attr}
\resizebox{\linewidth}{!}{%
\begin{tabular}{lcccccccccc}
\toprule
Method & Real & Big & Mid & Wuk & SD14 & SD15
& ADM & GLI & VQ & Avg. \\
\midrule
ResNet
& 16.1 & 10.6 & 56.3 & 27.6 & 12.1 & 22.6
& 17.1 & 20.1 & 10.1 & 21.4 \\
DIRE
& 22.6 & 11.6 & 0.0 & 26.6 & 22.1 & 20.6
& 18.1 & 22.6 & 10.6 & 17.2 \\
ESSP
& 17.6 & 13.6 & 49.7 & 27.1 & 19.6 & 17.6
& 16.6 & 27.1 & 13.1 & 22.4 \\
LIDA
& 83.4 & 98.5 & 69.3 & 13.1 & 23.6 & 50.3
& 47.2 & 55.3 & 45.7 & 54.0 \\
\rowcolor{blue!5}
\textbf{EvoKnow (router)}
& 98.5 & 93.1 & 29.5 & 35.7 & 2.1 & 65.0
& 63.5 & 68.0 & 38.1 & \textbf{54.8} \\
\bottomrule
\end{tabular}%
}
\end{table*}

This section analyzes the persistent state required for replay-free
analytical routing, its storage scaling, and its auxiliary
source-attribution behavior. EvoKnow is designed for binary real/fake
detection and residual knowledge retrieval rather than exact source
attribution; attribution is therefore reported only as a diagnostic.

\paragraph{Persistent State and Storage Scaling.}
In this work, ``replay-free'' specifically means that historical
AI-generated images, image features, and per-sample routing features are
not retained or replayed during incremental adaptation. This does not
imply zero persistent state. To continue routing updates without storing
historical generated images, the router retains the regularized Gram
inverse $\mathbf{R}$, the feature--target statistic $\mathbf{Q}$, and
router weights $\mathbf{W}=\mathbf{R}\mathbf{Q}$. Given
\begin{equation}
\mathbf{G}
=
\lambda_{\mathrm{RLS}}\mathbf{I}
+
\mathbf{X}^{\top}\mathbf{X},
\qquad
\mathbf{R}
=
\mathbf{G}^{-1},
\end{equation}
the recursive implementation maintains $\mathbf{R}$ directly. The
checkpoint additionally contains the fixed projection $\mathbf{P}$,
the mapping between router outputs and residual units, and update
metadata.

For routing dimension $m$ and $M$ incremental residual experts,
$\mathbf{R}$ requires $O(m^2)$ values, whereas $\mathbf{Q}$ and
$\mathbf{W}$ each require $O(mM)$. The projection requires $O(md)$
values, and expert parameters grow as $O(Mp)$. At $m=10{,}000$,
$\mathbf{R}$ occupies approximately 400 MB in FP32.
Table~\ref{tab:supp-router-state} summarizes the persistent router state
and its storage scaling. Thus, replay-free learning in EvoKnow avoids
retaining or replaying historical generated images and their
instance-level features, while retaining aggregated router statistics,
learned router parameters, and an expanding set of incremental residual
experts. It therefore avoids an image replay buffer but does not imply
zero persistent state.

\paragraph{Auxiliary Source-attribution Diagnostics.}
The following attribution diagnostic is conducted under
generator-based residual-unit organization, where each incremental residual expert has a one-to-one association with an observed generator domain. AIR
selects residual knowledge units rather than explicitly predicting exact
source identities. Since multiple generators can share related forensic
differences, source attribution is treated only as an auxiliary
diagnostic. In the generator-based setting, router outputs can be mapped
to source labels as
\begin{equation}
\hat c(x)=
\begin{cases}
\mathrm{Real}, & \hat y(x)=0,\\
g(r(x)), & \hat y(x)=1,
\end{cases}
\label{eq:supp-attribution}
\end{equation}
where $r(x)$ is the selected expert index, $\hat y(x)$ is the binary
prediction, and $g(\cdot)$ maps an expert to its associated generator.


Table~\ref{tab:supp-attr} reports auxiliary source-attribution results for the complete
detection-and-routing pipeline over one real-image class and eight
generator classes. These results complement the 14-class
frozen-representation probes in Table~\ref{tab:diagnostics}(b):
the probes assess the generator-discriminative information available
in the feature space, and the attribution diagnostic evaluates how
the deployed detection and routing outputs map to source labels.
The class-wise attribution results characterize the alignment between
expert selection and generator identity. In EvoKnow, expert retrieval
serves binary real/fake detection, and source attribution is reported
as an auxiliary diagnostic of the resulting routing behavior.

\end{document}